\documentclass{article}

\usepackage{microtype}
\usepackage{graphicx}
\usepackage{subcaption}
\usepackage{booktabs} 

\usepackage{algpseudocode}

\usepackage{hyperref}

\usepackage[accepted]{icml2023}

\usepackage{amsmath}
\usepackage{amssymb}
\usepackage{mathtools}
\usepackage{amsthm}

\usepackage[capitalize,noabbrev]{cleveref}

\theoremstyle{plain}

\theoremstyle{definition}

\theoremstyle{remark}

\usepackage[textsize=tiny]{todonotes}

\usepackage{amsmath,amsfonts,bm}

\def\eqref#1{equation~\ref{#1}}

\def\1{\bm{1}}

\def\rva{{\mathbf{a}}}

\def\rvs{{\mathbf{s}}}

\DeclareMathAlphabet{\mathsfit}{\encodingdefault}{\sfdefault}{m}{sl}
\SetMathAlphabet{\mathsfit}{bold}{\encodingdefault}{\sfdefault}{bx}{n}

\def\gD{{\mathcal{D}}}

\newcommand{\E}{\mathbb{E}}

\newcommand{\KL}{D_{\mathrm{KL}}}

\usepackage{multirow}     
\usepackage{enumitem}
\setlist[itemize]{noitemsep, topsep=0pt}
\setlist[enumerate]{noitemsep, topsep=0pt}

\definecolor{Green}{HTML}{09af00} 
\definecolor{Orange}{HTML}{fa8100} 
\newcommand{\currentTitle}{Mastering the Unsupervised Reinforcement Learning Benchmark from Pixels}

\definecolor{Data}{HTML}{398751}
\definecolor{Knowledge}{HTML}{1e558d}
\definecolor{Competence}{HTML}{bf3838}

\usepackage{stfloats}

\usepackage{pifont}
\newcommand{\cmark}{\ding{51}}%
\icmltitlerunning{ ~ \hfill \currentTitle \hfill } 

\begin{document}

\twocolumn[
\icmltitle{\currentTitle}



\icmlsetsymbol{equal}{*}

\begin{icmlauthorlist}
\icmlauthor{Sai Rajeswar}{equal,mila,snow}
\icmlauthor{Pietro Mazzaglia}{equal,ugent}
\icmlauthor{Tim Verbelen}{ugent}
\icmlauthor{Alexandre Piché}{snow} \\
\icmlauthor{Bart Dhoedt}{ugent}
\icmlauthor{Aaron Courville}{mila,cifar}
\icmlauthor{Alexandre Lacoste}{snow}
\end{icmlauthorlist}

\icmlaffiliation{mila}{Mila, Université de Montréal} 
\icmlaffiliation{snow}{ServiceNow Research}
\icmlaffiliation{ugent}{Ghent University - imec, Belgium}
\icmlaffiliation{cifar}{CIFAR Fellow}

\icmlcorrespondingauthor{Sai Rajeswar}{rajsai24@gmail.com}
\icmlcorrespondingauthor{Pietro Mazzaglia}{pietro.mazzaglia@ugent.be}

\icmlkeywords{Machine Learning, ICML}

\vskip 0.3in
]




\begin{table*}[b!]
\centering
\begin{tabular}{|c|c|c|c|c|c|c|}
\hline
\textbf{Approach}                      &  \textbf{Unsupervised PT} & \textbf{Model-based} & \textbf{Task-aware FT} & \textbf{Dyna-MPC} & \textbf{Performance (\%)} \\ \hline
DrQv2                           & & & & &  $17.66_{\pm1.09}$                         \\ \hline
DreamerV2                           & & \cmark & & &  $28.16_{\pm1.26}$                         \\ \hline
Disagreement & \cmark & & & &  $39.0_{\pm1.87}$                         \\ \hline
Plan2Explore (P2E)  & \cmark & \cmark & &                         & $76.59_{\pm 3.46}$                      \\ \hline
P2E + task-aware FT \textbf{\footnotesize{(ours)}} & \cmark & \cmark & \cmark &  &  $83.07_{\pm 2.01}$ \\ \hline  
P2E + Dyna-MPC \textbf{\footnotesize{(ours)}} & \cmark & \cmark & \cmark & \cmark     & $88.86_{\pm 1.76}$                      \\ \hline
\textbf{Ours} & \cmark & \cmark & \cmark & \cmark                         & $\mathbf{93.59_{\pm 0.84}}$                       \\ \hline
\end{tabular}
\caption{\textbf{Mastering URLB from pixels.} The table summarizes the results obtained by our method compared to previous approaches. We also show how the performance of Plan2Explore (P2E) \citep{sekar2020planning} increases when we combine it with our adaptation strategies.}
\label{tab:summary}
\end{table*}

\begin{abstract}
Controlling artificial agents from visual sensory data is an arduous task. Reinforcement learning (RL) algorithms can succeed but require large amounts of interactions between the agent and the environment. To alleviate the issue, unsupervised RL proposes to employ self-supervised interaction and learning, for adapting faster to future tasks. 
Yet, as shown in the Unsupervised RL Benchmark (URLB; \citet{laskin2021urlb}), whether current unsupervised strategies can improve generalization capabilities is still unclear, especially in visual control settings.
In this work, we study the URLB and propose a new method to solve it, using unsupervised model-based RL, for pre-training the agent, and a task-aware fine-tuning strategy combined with a new proposed hybrid planner, Dyna-MPC, to adapt the agent for downstream tasks. On URLB, our method obtains 93.59\% overall normalized performance, surpassing previous baselines by a staggering margin. The approach is empirically evaluated through a large-scale empirical study, which we use to validate our design choices and analyze our models. We also show robust performance on the Real-Word RL benchmark, hinting at resiliency to environment perturbations during adaptation.

\centering\textbf{Project website:} \\ \url{https://masteringurlb.github.io/} 

\end{abstract}

\section{Introduction}
\label{sec:introduction}

Modern successes of deep reinforcement learning (RL) have shown promising results for control problems \citep{levine2016, OpenAI2019Rubik, Lu2021AWOpt}. However, training an agent for each task individually requires a large amount of task-specific environment interactions, incurring huge redundancy and prolonged human supervision. 
Developing algorithms that can efficiently adapt and generalize to new tasks has hence become an active area of research. 


\printAffiliationsAndNotice{\icmlEqualContribution} 

In computer vision and natural language processing, unsupervised learning has enabled training models without supervision to reduce sample complexity on downstream tasks~\citep{Chen2020SimCLR, Radford2019GPT2}. In a similar fashion, unsupervised RL (URL) agents aim to learn about the environment without external reward functions, driven by intrinsic motivation~\citep{Pathak_curiosity, Burda19, bellemarecount}. The learned models can then be adapted to downstream tasks, aiming to reduce the required amount of interactions with the environment.

Recently, the Unsupervised RL Benchmark (URLB) \citep{laskin2021urlb} established a common protocol to compare self-supervised algorithms across several domains and tasks from the DMC Suite~\citep{tassa_dmcontrol}. In the benchmark, an agent is allowed a task-agnostic pre-training stage, where it can interact with the environment in an unsupervised manner, followed by a fine-tuning stage where, given a limited budget of interactions with the environment, the agent should quickly adapt for a specific task. 
However, the results obtained by \citet{laskin2021urlb} suggest that the benchmark is particularly challenging for current URL approaches, especially when the inputs of the agent are pixel-based images.

World models have proven highly effective for solving RL tasks from vision both in simulation \citep{Hafner2021DreamerV2, Hafner2019Dream} and in robotics \citep{Wu2022DayDreamer}, and they are generally data-efficient as they enable learning behavior in imagination \citep{Sutton1991Dyna}. Inspired by previous work on model-based exploration \citep{sekar2020planning} and by the idea that world models can efficiently leverage self-supervised data \citep{lecun2022path}, we adopted a world-model-based approach. 

In a preliminary large-scale study, we show that different URL strategies can be effectively combined with world-model-based agents, for unsupervised pre-training, leading to more solid performance in URLB from pixels, compared to model-free agents \citep{laskin2021urlb}. With our method, we further improve performance by leveraging a task-aware adaptation strategy and by introducing a new hybrid planner, Dyna-MPC, which allows exploiting the pre-trained world model even more, by enabling the agent to both learn and plan behavior in imagination. 


 Our contributions can be summarized as follow: 
 \begin{itemize}
     \item we perform a large-scale study on URLB showing that world models can be combined with different unsupervised RL approaches as an effective pre-training strategy for data-efficient visual control (Section \ref{sec:unsupervised_expl}),
     \item we introduce our method: an effective adaptation strategy that combines task-aware fine-tuning of the agent's components with a novel hybrid planner, \textit{Dyna-MPC}, enabling the agent to effectively combine behaviors learned in imagination with planning (Section \ref{subsec:plan}),
     \item we present state-of-the-art results on URLB from pixels, obtaining \textbf{93.59\% normalized performance}. We also extensively evaluate and analyze our method, to test its robustness to potential environment perturbations at adaptation time \citep{DulacArnold2020RWRL} and to understand current limitations (Section \ref{sec:result}).
 \end{itemize}

An extensive empirical evaluation, supported by more than 2k experiments, among main results, analysis and ablations, was used to carefully study URLB and analyse our method. We hope that our large-scale evaluation will inform future research towards developing and deploying pre-trained agents that can be adapted with considerably less data to more complex and realistic tasks, as it has happened with unsupervised pre-trained models for vision~\citep{Parisi2022PTVisionRobot} and language~\citep{Ahn2022DoAsICan}.


\section{Preliminaries} 

\textbf{Reinforcement learning.} The RL setting can be formalized as a Markov Decision Process (MDP), denoted with the tuple $\{\mathcal{S}, \mathcal{A}, T, R, \gamma\}$, where $\mathcal{S}$ is the set of states, $\mathcal{A}$ is the set of actions, $T$ is the state transition dynamics, $R$ is the reward function, and $\gamma$ is a discount factor.    
The objective of an RL agent is to maximize the expected discounted sum of rewards over time for a given task, also called return, and indicated as ${G_t = \sum_{k=t+1}^T \gamma^{(k-t-1)}r_k}$. In continuous-action settings, you can learn an actor, i.e. a model predicting the action to take from a certain state, and a critic, i.e. a model that estimates the expected value of the actor's actions over time. Actor-critic algorithms can be combined with the expressiveness of neural network models to solve complex continuous control tasks~\citep{haarnoja2018soft, LillicrapHPHETS15, Schulman2017PPO}.  

\textbf{Unsupervised RL.} In this work, we investigate the problem of fast adaptation for a downstream task, after a phase of unsupervised training and interaction with the environment. Our training routine, based on the setup of URLB~\citep{laskin2021urlb}, is made of two phases: a pre-training (PT) phase, where the agent can interact with a task-agnostic version of the environment for up to 2M frames, and a fine-tuning phase (FT), where the agent is given a task to solve and a limited budget of 100k frames.  During the PT phase, rewards are removed so that sensible information about the environment should be obtained by exploring the domain-dependent dynamics, which is expected to remain similar or unchanged in the downstream tasks. During FT, the agent receives task-specific rewards when interacting with the environment. As the agent has no prior knowledge of the task, it should both understand the task and solve it efficiently, in a limited interaction budget.
The URL benchmark consists of three control domains, Walker, Quadruped and Jaco, and twelve tasks, four per domain. To evaluate the agents, snapshots of the agent are taken at different times during training, i.e. 100k, 500k, 1M, and 2M frames, and fine-tuned for 100k frames. Returns are normalized using results from a supervised baseline (see Appendix \ref{app:reference_scores}).


\begin{figure*}[tb!]
    \centering
    \begin{minipage}[t]{\linewidth}
        \centering
        \includegraphics[width=\linewidth]{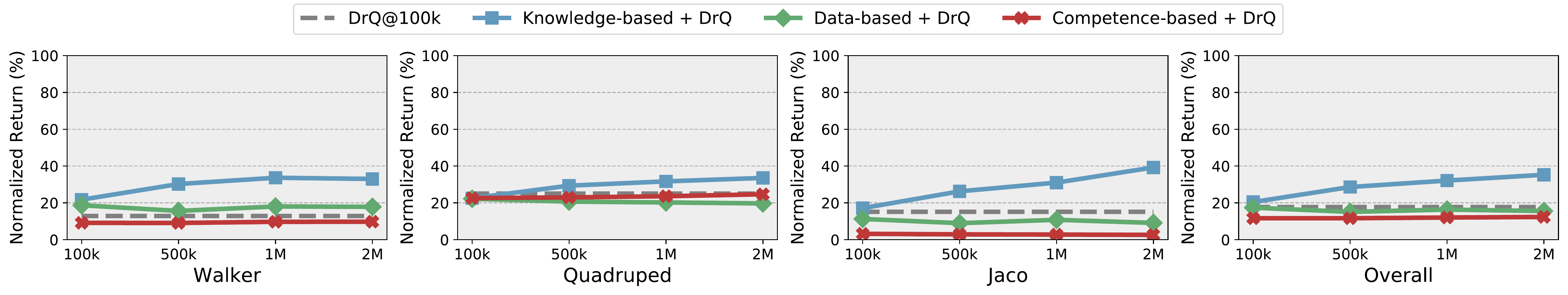}
        \subcaption{Aggregated results from \citet{laskin2021urlb} of unsupervised RL to pre-train the model-free DrQ agent.}
        \label{subfig:urlb}
    \end{minipage}
    \begin{minipage}[t]{\linewidth}
        \centering
        \includegraphics[width=\linewidth]{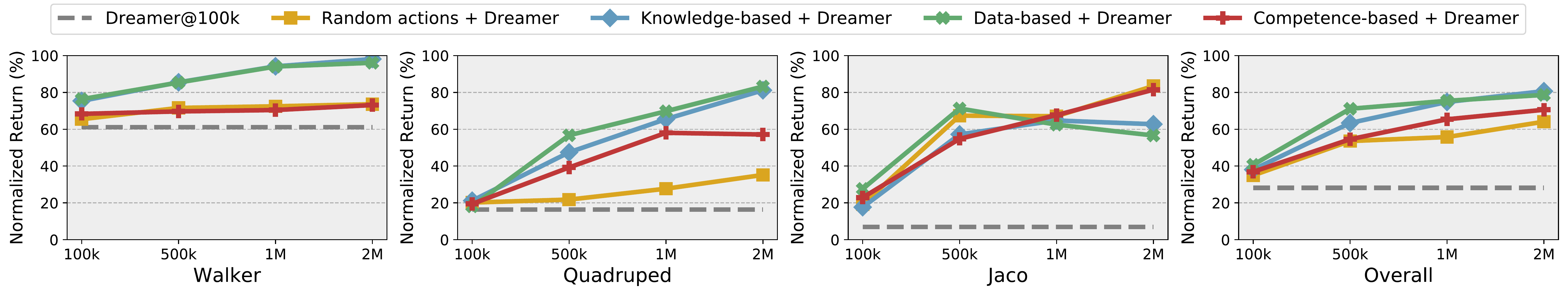}
        \subcaption{Our aggregated results of unsupervised RL to pre-train the model-based Dreamer agent.}
        \label{subfig:ours}
    \end{minipage}
    \caption{\textbf{Unsupervised pre-training.} Aggregated performance of different URL techniques for PT, with FT snapshots taken at different times along training. (a) With the model-free DrQ agent, performance slightly improves over time only using knowledge-based techniques. (b) With the model-based Dreamer agent, performance is higher and, overall, improves for all techniques. We also report Dreamer@100k and DrQ@100k results, which are obtained in 100k FT steps with no PT.}
    \label{fig:expl_results}
\end{figure*}

\textbf{World models.} In this work, we ground upon the DreamerV2 agent~\citep{Hafner2021DreamerV2}, which
 learns a world model~\citep{worldmodel, hafner2019planet} predicting the outcomes of actions in the environment. The dynamics is captured into a latent space $\mathcal{Z}$, providing a compact representation of the high-dimensional inputs. 

 \begin{minipage}{\columnwidth}
The world model consists of the following components:
\begin{center}
\begin{tabular}{llll}
    Encoder: & $e_t = f_\phi(s_t)$, \\ 
    Decoder:  & $p_\phi(s_t|z_t)$, \\ 
    Dynamics: & $p_\phi(z_t|z_{t-1}, a_{t-1})$, \\
    Posterior: & $q_\phi(z_t|z_{t-1}, a_{t-1}, e_t)$.
\end{tabular}
\end{center}
\end{minipage}

The model states $z_t$ have both a deterministic component, modeled using the recurrent state of a GRU~\citep{gru14}, and a (discrete) stochastic component. The encoder and decoder are convolutional neural networks (CNNs) and the remaining components are multi-layer perceptrons (MLPs). The world model is trained end-to-end by optimizing an evidence lower bound (ELBO) on the log-likelihood of the data collected in the environment~\citep{hafner2019planet, Hafner2019Dream}. For the encoder and the decoder networks, we used the same architecture as in \citet{Hafner2021DreamerV2}.

For control, the agent learns latent actor $\pi_\theta(a_t|z_t)$ and critic $v_\psi(z_t)$ networks.
Both components are trained online within the world model, by imagining the model state outcomes of the actions produced by the actor, using the model dynamics. Rewards for imagined trajectories are provided by a reward predictor, $p_\phi(r_t|z_t)$ trained to predict environment rewards, and they are combined with the critic predictions to produce a GAE-$\lambda$ estimate of the returns~\citep{Schulman2015GAE}. The actor maximizes these returns, backpropagating gradients through the model dynamics. 
The hyperparameters for the agent, which we keep fixed across all domains and tasks, can be found in Appendix \ref{app:hyperparams}. 


\section{Unsupervised Model-based Pre-training}
\label{sec:unsupervised_expl}




In the PT stage, unsupervised RL can be used to explore the environment, collecting the data to train the components of the agent. The resulting networks are then used to initialize respective components in the agent deployed for the downstream task, aiming to reduce sample complexity. 



Unsupervised RL methods can be grouped into three categories~\citep{laskin2021urlb}: knowledge-based, which aim to increase the agent's knowledge by maximizing error prediction \citep{Pathak_curiosity, pathak19a_disagreement, rnd}, data-based, which aim to achieve diversity of data \citep{yarats2021proto, liu2021unsupervised_apt} and competence-based, which aim to learn diverse skills \citep{aps, eysenbach2018diversity}. In Figure \ref{subfig:urlb} we report the results from \citet{laskin2021urlb}, showing that none of these approaches is particularly effective on URLB from pixels when combined with the DrQv2 model-free agent \citep{Yarats2021DrQ-v2}, state-of-the-art in RL from pixels, where the data collected with unsupervised RL is used to pre-train the agent's actor, critic, and encoder. The cause of this underwhelming performance is that all the pre-trained components in model-free agents rely on the reward function to maximize. As rewards during the PT stage are intrinsic rewards coming from unsupervised RL, they miss capturing important aspects of the environment, such as the dynamics of the environment, that could be useful to efficiently adapt to downstream tasks. 

World model-based agents, instead, can be used to effectively exploit unsupervised data collection, as they focus on learning the dynamics of the environment and then leverage the learned model to learn actions in latent imagination. To strengthen this thesis, we perform a large-scale study, including multiple unsupervised RL approaches and using them to pre-train the Dreamer's agent components. As knowledge-based methods we employ ICM~\citep{Pathak_curiosity}, LBS~\citep{Mazzaglia2021SelfSupervisedEV}, Plan2Explore (P2E; \citep{sekar2020planning}), and RND~\citep{rnd}. As a data-based approach, we choose APT~\citep{liu2021unsupervised_apt}, and as competence-based approaches, we adopt DIAYN~\citep{eysenbach2018diversity} and APS~\citep{aps}. Finally, we also test random actions, as a naive maximum entropy baseline~\citep{haarnoja2018soft}.


\begin{figure*}[t!]
    \centering
    \includegraphics[width=\textwidth]{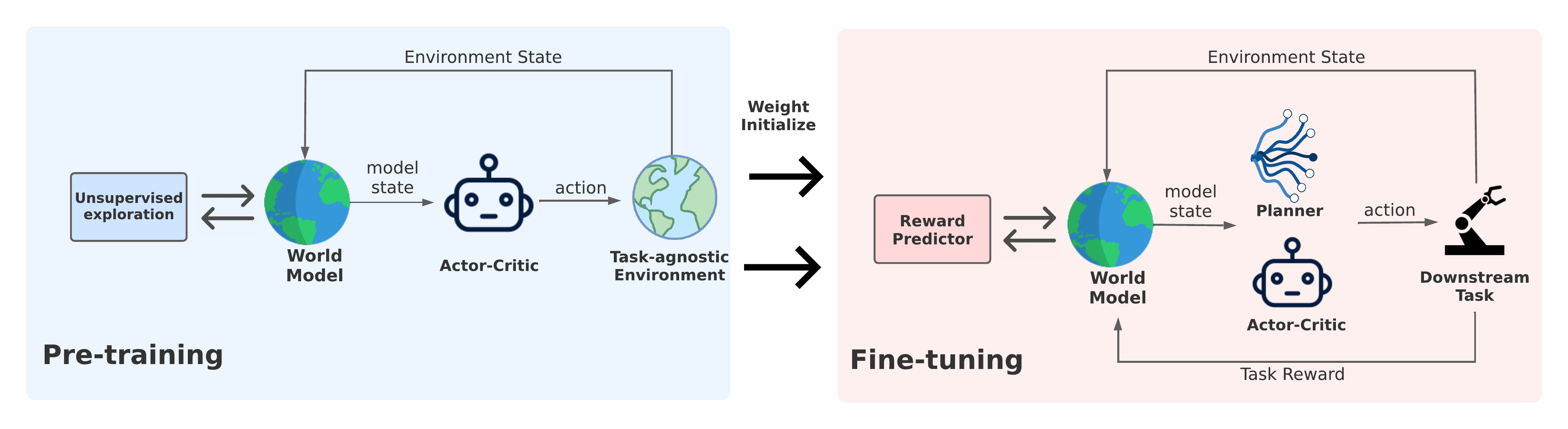}
    \caption{\textbf{Method overview.} Our method considers a pre-training (PT) and a fine-tuning (FT) stage. During pre-training, the agent interacts with the environment through unsupervised RL, maximizing an intrinsic reward function, and concurrently training a world model on the data collected. During fine-tuning, the agent exploits some of the pre-trained components and plans in imagination, to efficiently adapt to different downstream tasks, maximizing the rewards received from the environment.}
    \label{fig:overview}
\end{figure*}

Aggregating results per category, in Figure \ref{subfig:ours}, we show that by leveraging a pre-trained world model the overall performance improves over time for all categories, as opposed to the model-free results, where only knowledge-based approaches slightly improve. In particular, data-based and knowledge-based methods are more effective in the Walker and Quadruped domains, and random actions and competence-based are more effective in the Jaco domain. Detailed results for each method, which are available in Appendix \ref{app:add_results}, also show that, in contrast with the findings in \citet{sekar2020planning}, many unsupervised RL approaches can be combined with world models for efficient exploration. This merit could be attributed to the way we carefully adapted these methods to effectively work with world models'.
Details on the implementation are provided in Appendix~\ref{app:exploration} and the code is available on the \href{https://masteringurlb.github.io/}{project website}.



\section{Method}
\label{subsec:plan}

In the previous section's large-scale study on URLB, we showed that learning a model-based agent with data collected using unsupervised RL constitutes an effective pre-training strategy. Based on that, we focus our method on efficiently adapting the pre-trained world-model-based agents for downstream tasks. Our approach can be summarized as:
\begin{itemize}
\item employing a task-aware FT strategy, which only adapts the PT agent's modules that we expect to be sensible for the downstream task;
\item adopting a hybrid planner, which allows to further exploit the PT world model by learning and planning in imagination. For this purpose, we propose a new algorithm: \emph{Dyna-MPC}.
\end{itemize}
An overview of the method is illustrated in Figure~\ref{fig:overview} and the algorithm is presented in Appendix~\ref{app:algorithm}.

\textbf{Task-aware fine-tuning.} In the context of URLB, where the environment dynamics is unchanged between the PT and FT stage, some of the components learned during unsupervised interaction, such as the world model, can be reused for fast adaptation during FT. However, as the reward is changing from pseudo-reward to task reward when changing from the PT to the FT phase, it is not clear if pre-training of the actor and critic can help the downstream task, a factor which was not accounted for in previous work \citep{laskin2021urlb, sekar2020planning} and in the results in Section \ref{sec:unsupervised_expl}. 

The actor's actions during the unsupervised stage drive the agent to explore new transitions or to reach unseen states of the environment. When moving to the FT stage, these actions can be useful to quickly explore some environment transitions in dense reward tasks, like the Quadruped and Walker ones in URLB, where the agent is rewarded for each state change. However, in sparser reward settings, like the Jaco tasks, the agent's actions need to find the rewards corresponding to some specific areas of the environment. In these cases, actions that repeatedly drive the agent far from the starting state, potentially missing the task target, may actually make the adaptation stage more difficult. 

The critic's predictions, trained on the intrinsic rewards during the unsupervised interaction phase, would hardly transfer to a specific downstream task reward function, given the difference in reward scale and value. While we believe that finding ways to adapt the critic's prediction  might be interesting, e.g. by re-scaling them to be closer to the downstream task returns, we choose to discard the pre-trained critic and re-learn it from scratch during fine-tuning.

To summarize, moving our agent to the adaptation stage, we do always keep and fine-tune the PT model and we do always discard the PT critic. As for the PT actor, this is fine-tuned when the reward is dense, e.g. Walker and Quadruped tasks, but we discard it in sparse reward tasks, e.g. Jaco.


\textbf{Learning and planning in imagination.} Knowing a model of the environment, traditional model-based control approaches, e.g. model predictive control (MPC)~\citep{Williams2015MPPI, Chua2018PETS_MPC, Richards2005RobustMPC}, can be used to plan the agent's action. Nonetheless, using actor-critic methods has several advantages, such as amortizing the cost of planning by caching previously computed (sub)optimal actions and computing long-term returns from a certain state, without having to predict outcomes that are far in the future. More recent hybrid strategies, such as LOOP \citep{Silkchi2020LOOP} and TD-MPC \citep{Hansen2022TD-MPC}, allow combining the actor's predictions with trajectories sampled from a distribution over actions that is iteratively improved \citep{rubinstein2004cross}. 

 As in URLB we pre-train a world model, we could exploit planning in latent space to adapt with limited additional environment interaction. One problem with the above strategies is that they are based upon learning off-policy actor and critic, which in our context would prevent us from exploiting the PT model to learn the actor and critic in imagination. In order to enable hybrid planning with the behavior learned in imagination \citep{Hafner2019Dream}, we develop a new approach, which we call \emph{Dyna-MPC}, that combines the actor and critic learned in imagination with an MPPI-like sampling strategy \citep{Williams2015MPPI} for planning. 
 

\subsection{Dyna-MPC}


As detailed in Algorithm \ref{alg:inference}, at each time step, we imagine a set of latent trajectories using the model, by sampling actions from a time-dependent multivariate gaussian and from the latent actor policy, trained in imagination \citep{Hafner2019Dream}. Returns are estimated using reward predictions by the model and the critic. An iterative strategy \citep{Williams2015MPPI} is used to update the parameters of the multivariate gaussian for $J$ iterations. 
One significant difference with previous approaches is that the policy in Dyna-MPC is learned on-policy in imagination, thus no correction for learning off-policy is required \citep{Silkchi2020LOOP}. 

The critic is learned in the model's imagination, computing the expected value of the actor's actions using GAE-$\lambda$ estimates of the returns \citep{Schulman2015GAE}:
\begin{equation}
\label{eq:dream-return}
V^\lambda_t =
r_t + \gamma_t
\begin{cases}
  (1 - \lambda) v_\psi(z_{t+1}) + \lambda V^\lambda_{t+1} & \text{if}\quad t<H, \\
  v_\psi(z_H) & \text{if}\quad t=H, \\
\end{cases}
\end{equation}
where $r_t$ is the reward for state $z_t$, yielded by the reward predictor of the world model, and $H$ is the imagination horizon. 
When computing returns for the action's iterative update procedure we use the same return estimates.

At each step, we iteratively fit the parameters of a time-dependent multivariate Gaussian distribution with diagonal covariance, updating mean and standard deviation parameters using an importance-weighted average of the top-k trajectories with the highest estimated returns. At every step, $N$ trajectories $\Pi_i = \{ a_{0,i}, a_{1,i}, ..., a_{H,i} \}$ of length $H$ are obtained sampling actions from the distributions $a_t \sim \mathcal{N}(\mu_t, \sigma_t^{2}\mathrm{I})$ and $N_\pi$ trajectories are sampled from the actor network $a_t \sim \pi_\theta(a_t|z_t)$ and their outcomes are predicted using the model. 
At each iteration, first, the top-k trajectories with the highest returns are selected, then the distribution parameters are updated as follows:
\begin{align}
    \label{eq:mppi-return-norm}
    &\mu = \textstyle\sum_{i=1}^{k} \rho_i \Pi_{i}^{\star}\,,~
    \sigma = \max (\sqrt{ \textstyle\sum_{i=1}^{k} \rho_{i} (\Pi_{i}^{\star} - \mu)^{2}} \,,~\epsilon)\,, 
\end{align} 
where $\rho_{i} = \exp(\tau V^\lambda_{i}) / \sum_j \exp(\tau V^\lambda_{j})$, $\tau$ is a temperature parameter, $\star$ indicates the trajectory is in the top-k, and $\epsilon$ is a clipping factor to avoid too small standard deviations \citep{Hansen2022TD-MPC}. To reduce the number of iterations required for convergence, we reuse the 1-step shifted mean obtained at the previous timestep \citep{Argenson2020MOPO}.

\begin{algorithm}[t]
\caption{~~Dyna-MPC}
\label{alg:inference}
\begin{algorithmic}[1]
\Require Actor $\theta$, Critic $\psi$, World Model $\phi$ \\
~~~~~~~~~~$\mu, \sigma$: initial parameters for sampling actions\\
~~~~~~~~~~$N, N_{\pi}$: num trajectories, num policy trajectories\\
~~~~~~~~~~$\mathbf{z}_{t}, H$: current model state, planning horizon
\For{each iteration $j=1..J$}
\State Sample $N$ trajectories of length $H$ from $\mathcal{N}(\mu, \sigma^{2} \mathrm{I})$, starting from $z_t$
\State Sample $N_{\pi}$ trajectories of length $H$ using the actor $\pi_{\theta}$, starting from $z_t$
\State Predict future states using the model and expected returns using reward and critic predictions (Eq. \ref{eq:dream-return}) 
\State Update $\mu$ and $\sigma$ (Eq. \ref{eq:mppi-return-norm})
\EndFor
\State \textbf{return} $\mathbf{a_t} \sim \mathcal{N}(\mu_t, \sigma_t^{2} \mathrm{I})$
\end{algorithmic}
\end{algorithm}

\section{Evaluation and Analysis}
\label{sec:result}

For all experiments, results are presented with at least three random seeds. 

\begin{figure*}[t]
    \centering
    \includegraphics[width=0.9\linewidth]{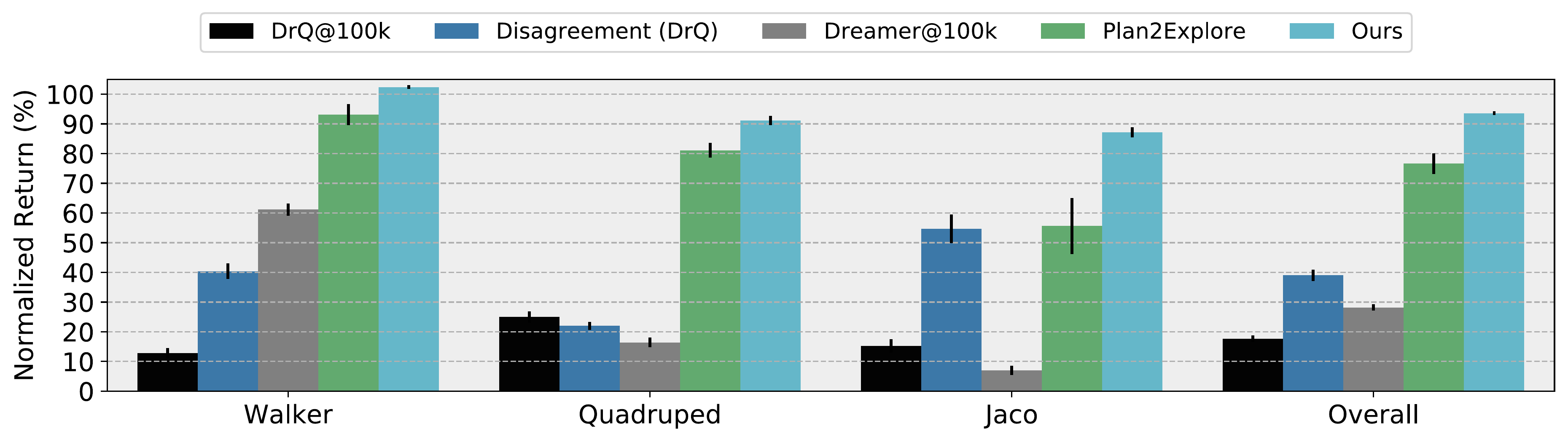}
    \caption{\textbf{URL Benchmark.} Our method obtains the highest overall performance on URLB, largely outperforming previous approaches.}
    \label{fig:first_page}
\end{figure*}
\begin{figure*}[b]
\begin{minipage}{.48\textwidth}
\centering
    \includegraphics[width=\columnwidth]{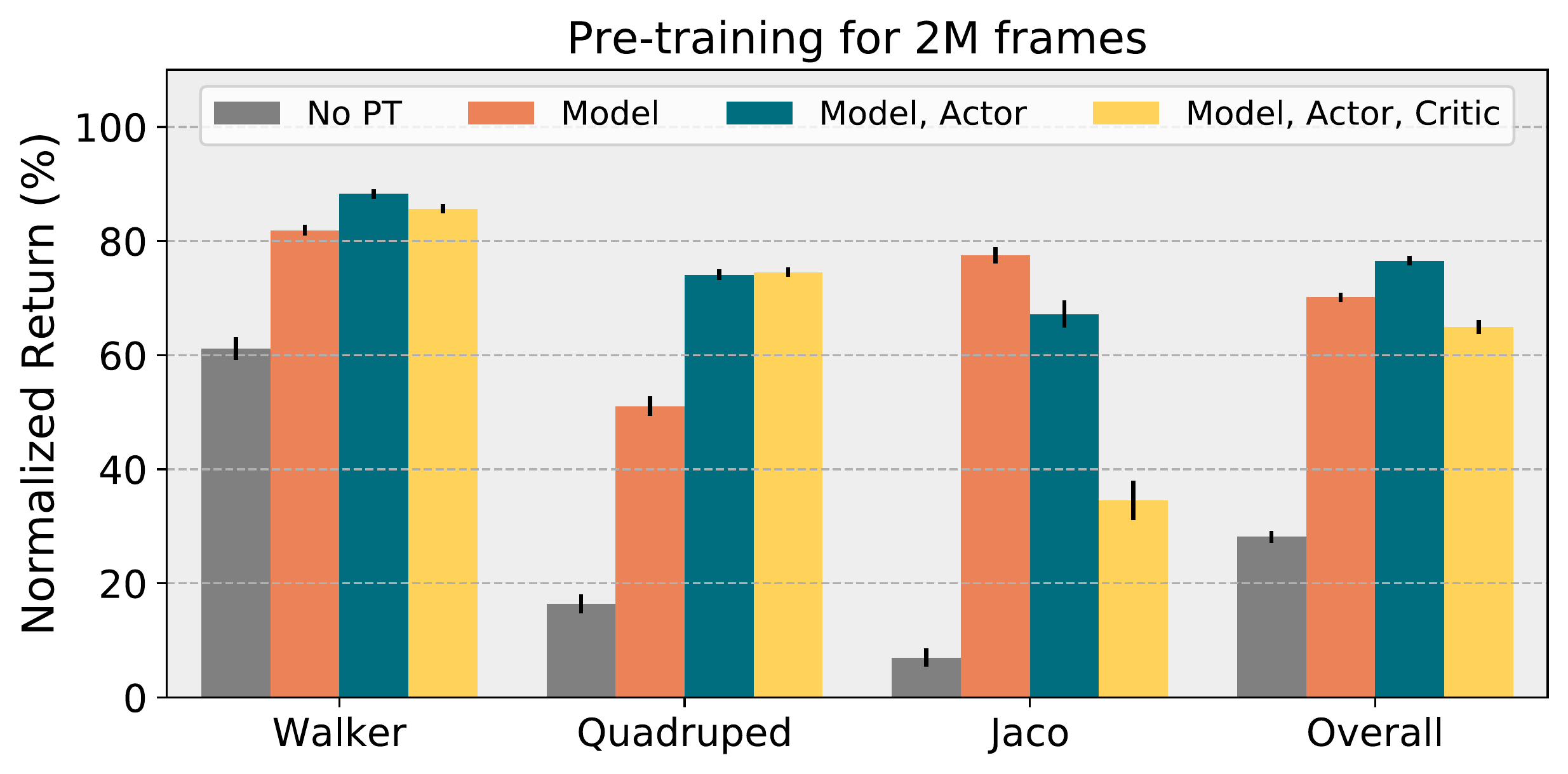}
    \caption{\textbf{Task-aware fine-tuning.} Effects of fine-tuning different sets of pre-trained components of the agent.}
    \label{fig:modules_results}
\end{minipage}
\hfill
\begin{minipage}{.48\textwidth}
    \centering
    \includegraphics[width=\columnwidth]{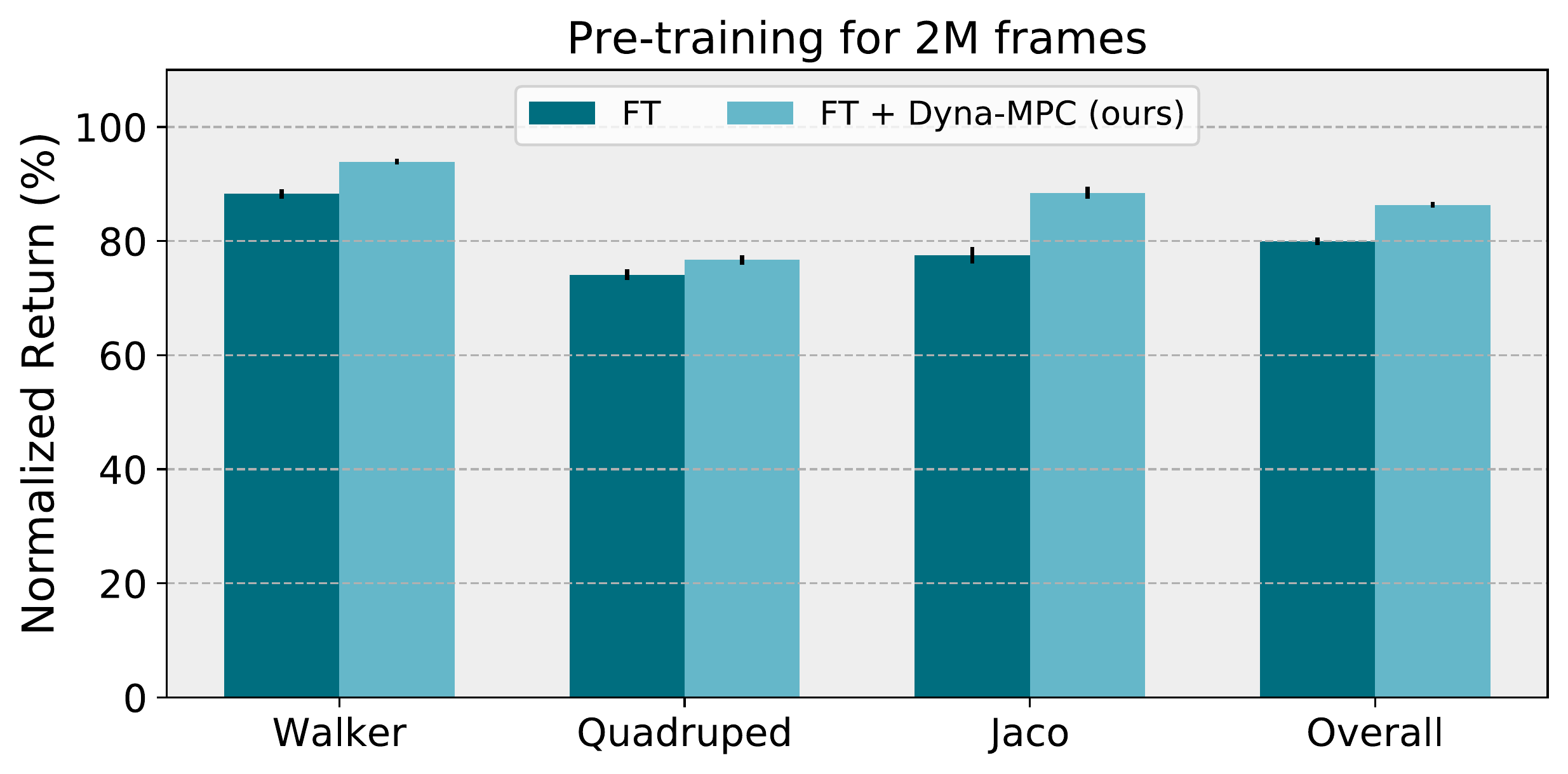}
    \caption{\textbf{Dyna-MPC.} Using Dyna-MPC during the adaptation stage improves performance in all domains.}
    \label{fig:planning_improv}
    \end{minipage}
\end{figure*}

\subsection{Unsupervised Reinforcement Learning Benchmark}
\label{subsec:ft}

In Figure \ref{fig:first_page}, we compare the results of Disagreement, the best-performing algorithm from the URLB paper \citep{laskin2021urlb}, and of Plan2Explore \citep{sekar2020planning} with our approach. We also report the scores of DrQv2 and DreamerV2 after 100k FT frames, with no pre-training. The performance of our method is superior in all domains. With respect to Disagreement, we improve performance by a staggering ~55\% margin. With respect to Plan2Explore, we improve performance by ~17\%. We highlight that the main differences with Plan2Explore are: (i) we employ LBS for unsupervised data collection \citep{Mazzaglia2021SelfSupervisedEV}, as this showed to be performing better in some domains (see Appendix \ref{app:add_results}), (ii) we employ task-aware FT, (iii) we adopt Dyna-MPC. 

As also reported in Table \ref{tab:summary}, Plan2Explore performance can also improve significantly when combined with our introduced adaptation strategies. We further validate these strategies through ablations in the following paragraphs.


\textbf{Task-aware fine-tuning.} We test different fine-tuning configurations, where we copy the weights of some of the PT components into the agent to fine-tune for the downstream task. To increase the generality of this ablation, we run the tests for all the unsupervised RL methods that we presented in Section \ref{sec:unsupervised_expl} and show aggregated results in Figure \ref{fig:modules_results} (detailed results per each method in Appendix \ref{app:add_results}).

Overall, fine-tuning the PT world model provides the most significant boost in performance, strengthening the hypothesis that world models are very effective with unsupervised RL. As we expected, using a PT critic is systematically worse and this can be explained by the discrepancy between intrinsic rewards and task rewards. Finally, fine-tuning the actor improves performance slightly in Walker tasks and remarkably in Quadruped tasks, which are the dense reward tasks, but it is harmful in the Jaco sparse reward tasks. 


\textbf{Dyna-MPC.} We use the world models and actors pre-trained with all the different unsupervised strategies we considered (see Section \ref{sec:unsupervised_expl}) and test their FT performance with and without planning with Dyna-MPC. Aggregated scores are reported in Figure \ref{fig:planning_improv}, and detailed results for each method are available in Appendix \ref{app:add_results}. We observe that adopting Dyna-MPC is always beneficial, as it improves the average performance in all domains.



\subsection{Real-World Reinforcement Learning Benchmark}
\label{subsec:rwrl}


\begin{figure}[t]
    \centering
    \includegraphics[width=\linewidth]{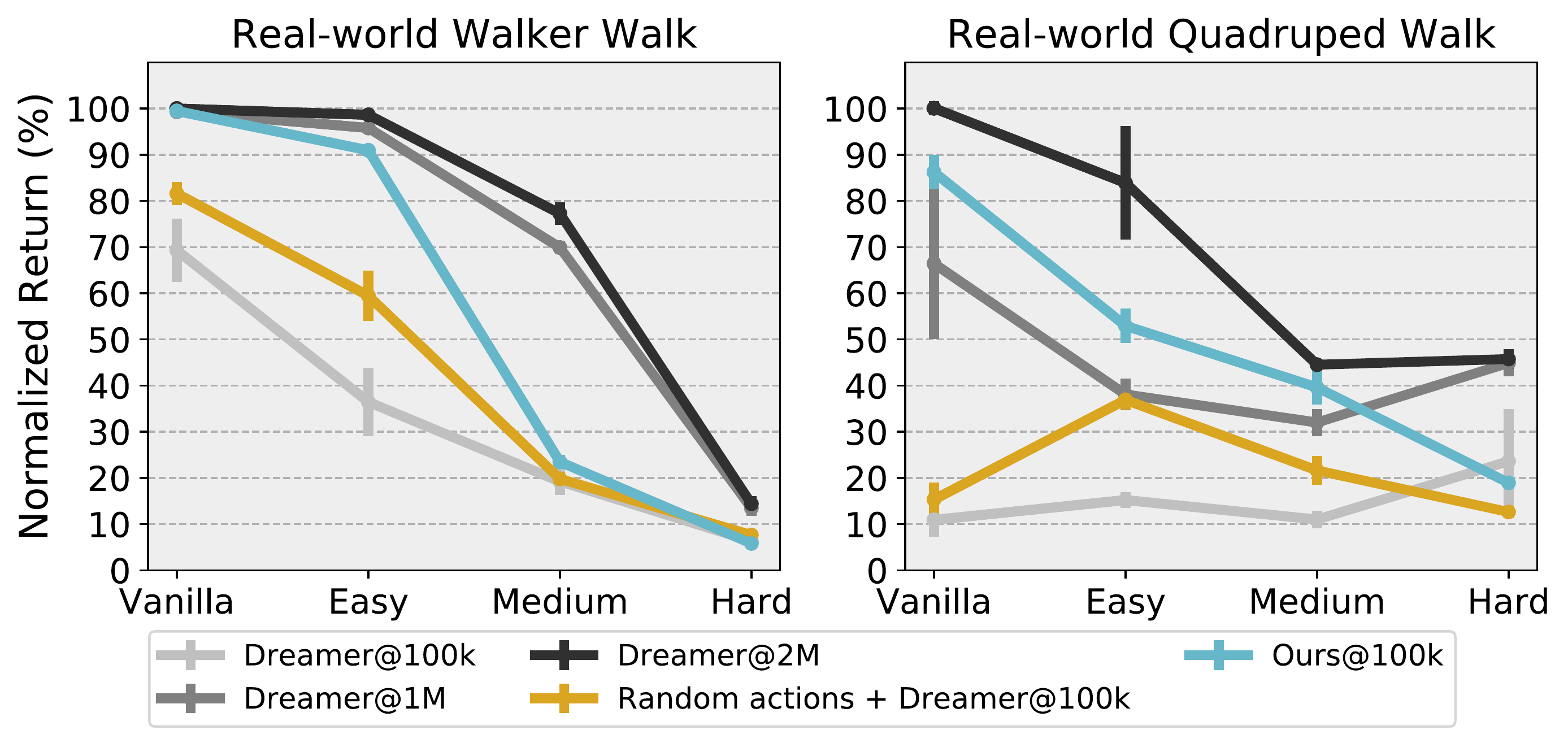}
    \caption{\textbf{RWRL tasks.} We evaluate our method and the baselines on the perturbated environments from the RWRL suite.}
    \label{fig:rwrl}
\end{figure}

We employ vision-based variants of the Walker Walk and Quadruped Walk tasks from the RWRL suite \citep{DulacArnold2020RWRL}  to test the robustness of our method. These tasks introduce system delays, stochasticity, and perturbations of the robot's model and sensors, which are applied with three degrees of intensity to the original environment, i.e. `easy', `medium', and `hard' (details in Appendix \ref{app:rwrl}).

In Figure \ref{fig:rwrl}, we present the results of our approach, pre-trained on the non-perturbed environment and fine-tuned on the environment wih perturbations, and compare it to training Dreamer from scratch on the perturbed environment for 100k, 1M, and 2M frames. We also add a random actions + Dreamer baseline, also pre-trained on the non-perturbed environment, to see whether our approach outperforms pre-training on randomly collected data. 

Overall, we found that fine-tuning PT models offer an advantage over training from scratch for 100k frames, despite all the variations in the environment. Furthermore, on the Quadruped Easy and Medium settings, our method performs better than Dreamer@1M and not far from Dreamer@2M while using 10x and 20x less task-specific data, respectively. Our method also performs close to Dreamer@1M/2M in the Walker Easy task. Finally, our method also strongly outperforms random actions in the `easy' and `medium' settings, showing that a better PT model yields higher FT performance, even when the dynamics of the downstream task is affected by misspecifications and noisy factors.





\subsection{Extended Analysis}
\label{subsec:analysis}

To better analyze the learned components, we conducted a range of additional studies. For conciseness, detailed descriptions of the experimental settings are deferred to Appendix~\ref{app:additional_analysis} and here we briefly summarize the takeaways. 

\begin{figure}[t]
    \centering
    \includegraphics[width=\columnwidth]{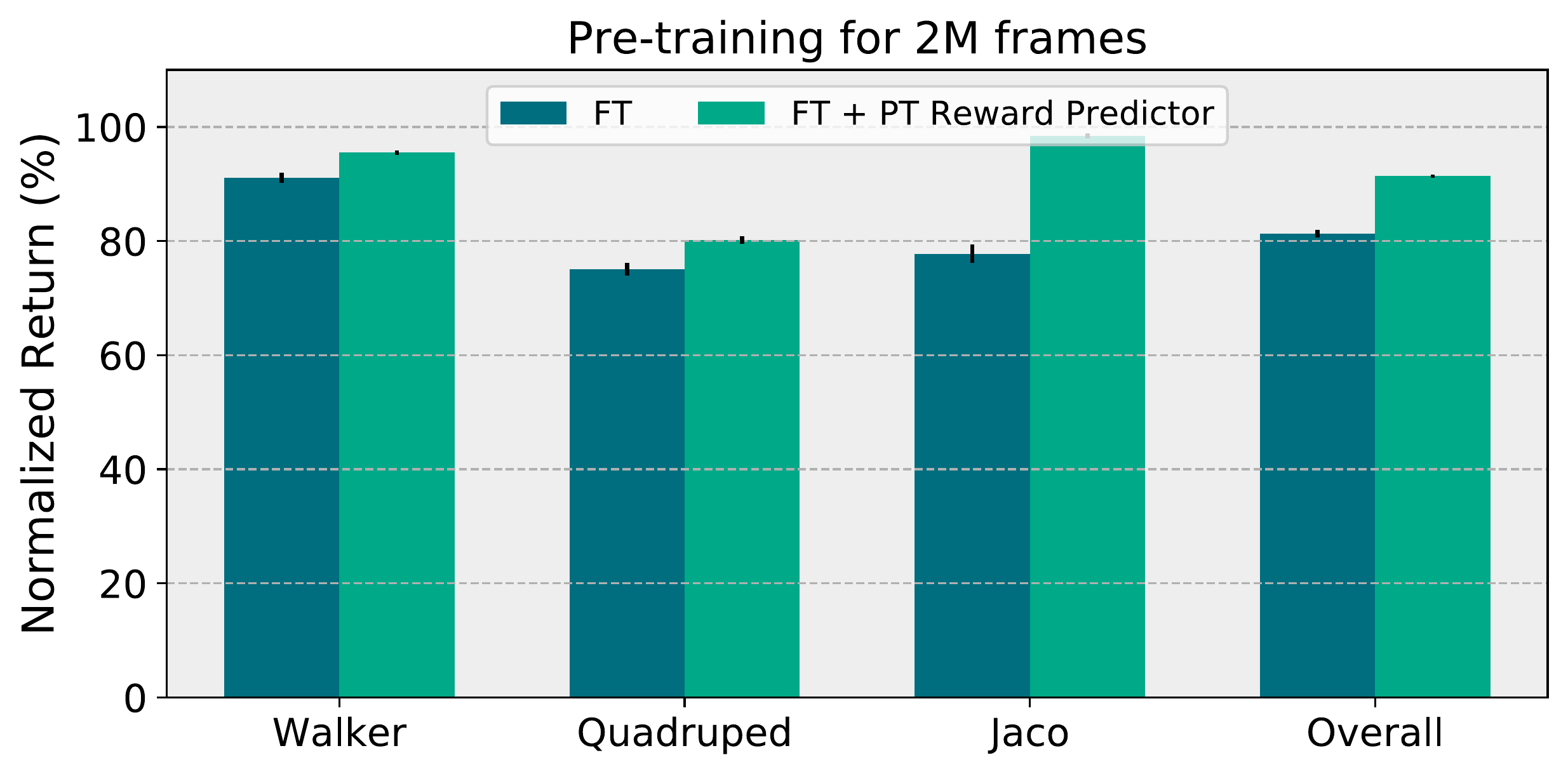}
    \caption{Ablation study about knowing the task from the PT stage.} 
    \label{fig:reward_ablation}
\end{figure}

\paragraph{Learning rewards online.} We verify whether having to discover and learn the reward function during FT impacts performance. In Figure \ref{fig:reward_ablation}, we compare against agents that (violating the URLB settings) know the task in advance and can pre-train a reward predictor during the PT stage. We see that learning the reward predictor does not affect performance significantly for dense-reward tasks, such as the Walker and Quadruped tasks. However, in sparser reward tasks, i.e. the Jaco ones, knowing reward information in advance provides an advantage. Efficient strategies to find sparse rewards efficiently represent a challenge for future research. More details in Appendix \ref{app:learning-reward-predictor}.

\begin{figure}[t]
    \centering
    \includegraphics[width=\columnwidth]{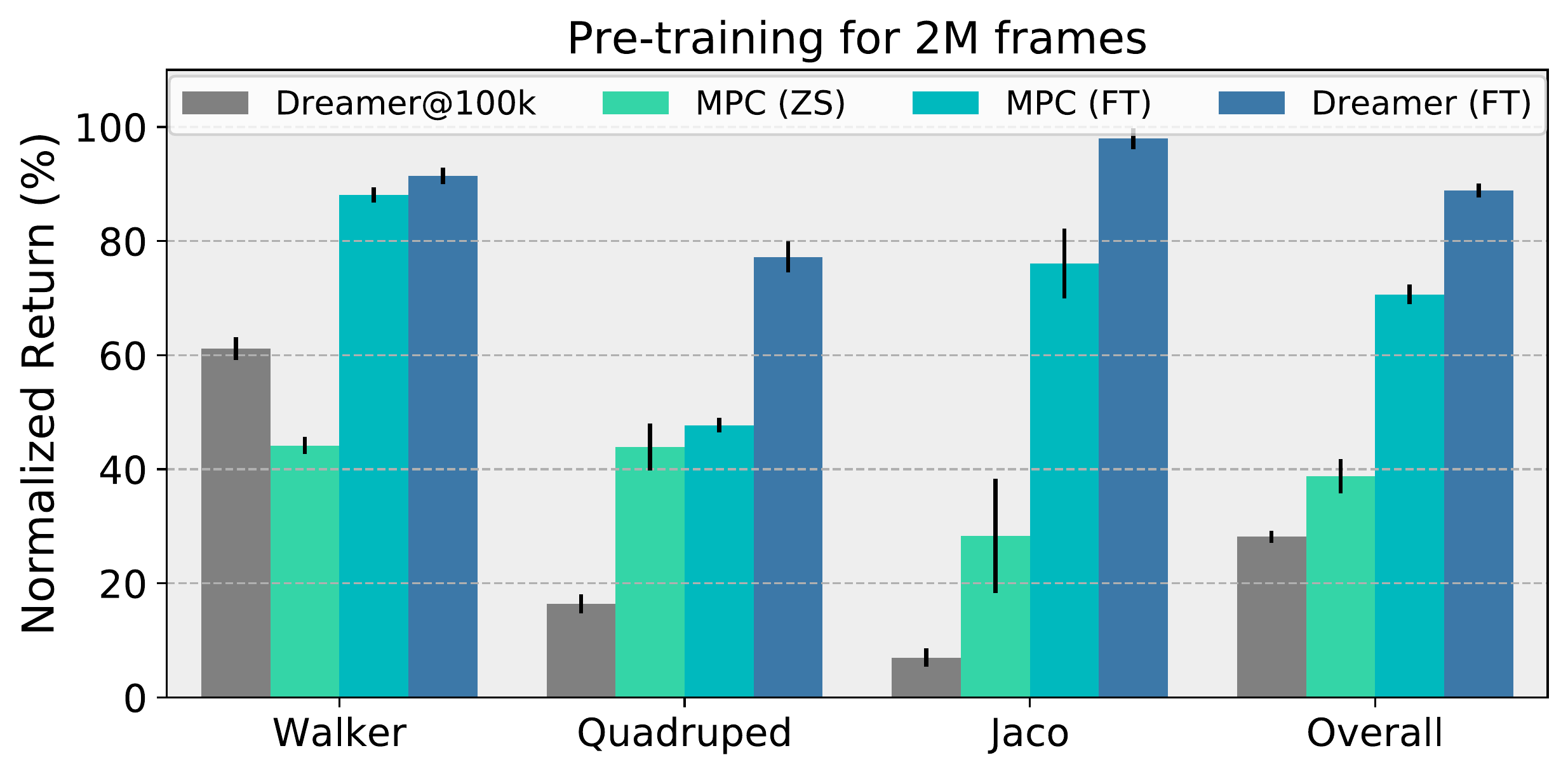}
    \caption{Zero-shot (ZS) vs fine-tuned (FT) performance. } 
    \label{fig:mpc_results}
\end{figure}

\paragraph{Zero-shot adaptation.} Knowing a reward predictor from PT, it could be possible to perform zero-shot control with MPC methods if the model and the reward function allow it. In Figure \ref{fig:mpc_results}, we show that despite the zero-shot MPC (ZS) offers an advantage over Dreamer@100k, the FT phase is crucial to deliver high performance on the downstream tasks, as the agent uses this phase to collect missing information about the environment and the task. Further details in Appendix \ref{app:zero-shot-perf}. 

\begin{figure}[t!]
    \centering
    \includegraphics[width=\columnwidth]{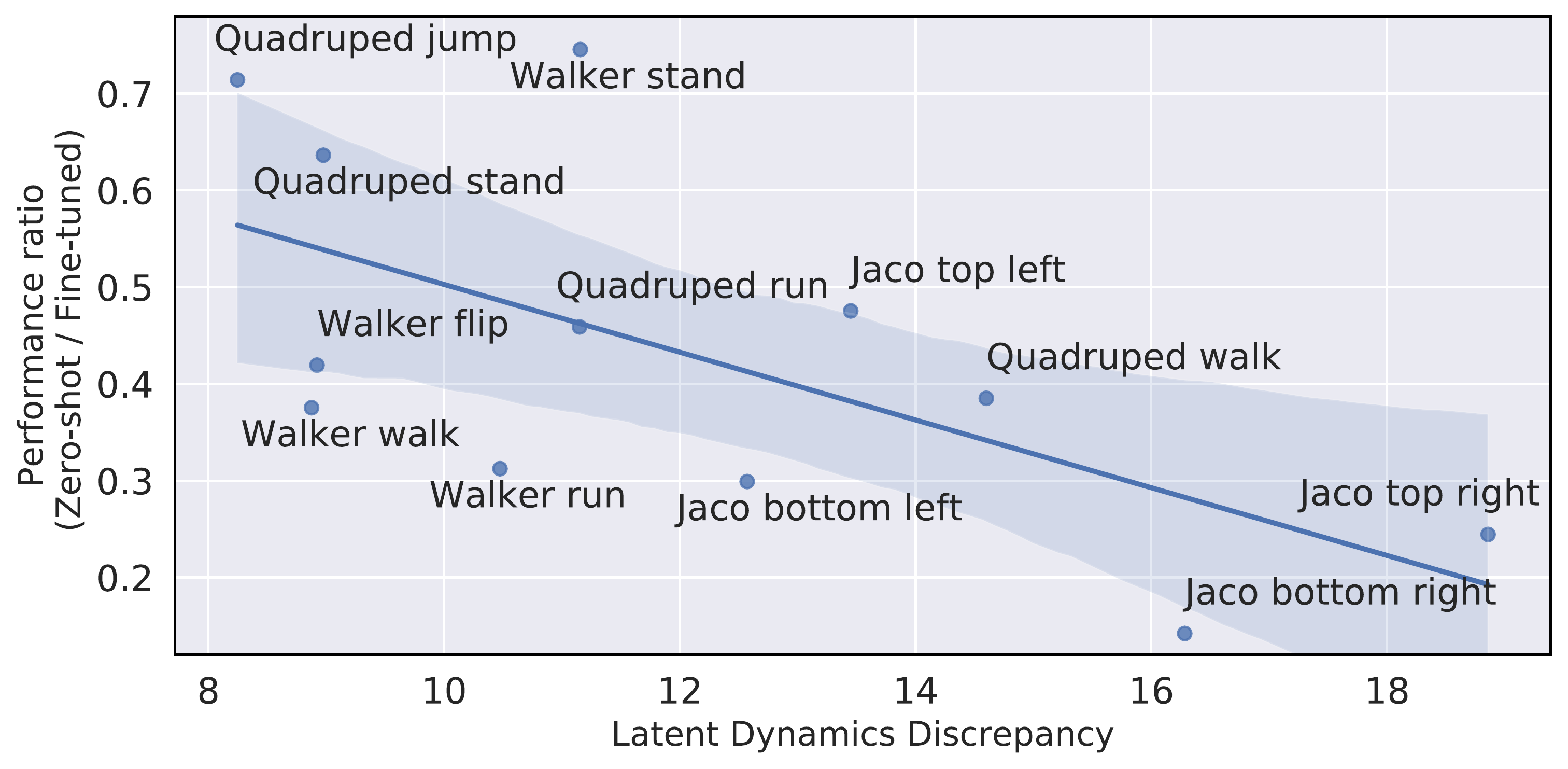}
    \caption{LDD and performance correlation. The line and shade represent a linear regression model fit and its confidence intervals.}
    \label{fig:model_miss}
\end{figure}

\paragraph{Latent dynamics discrepancy.} 
We propose a novel metric, \emph{Latent Dynamics Discrepancy} (LDD), evaluating the distance between the latent predictions of the PT model and the same model after FT on a task. In Figure \ref{fig:model_miss}, we show the correlation between LDD and the performance ratio between using the PT model and the FT model for planning (see Appendix~\ref{app:LD-discrepancy} for a detailed explanation).
We observed a strong negative Pearson correlation ($-0.62$, p-value: $0.03$), highlighting that updates in the model dynamics during FT played a significant role in improving performance. 

\textbf{Unsupervised rewards and performance.}
We analyze the correlation between the performance of different agents and their intrinsic rewards for optimal trajectories obtained by an oracle agent in Table \ref{tab:novelty}. In particular, the correlation for LBS, which overall performs best in URLB, has a statistical significance, as its p-value is $<0.05$. We believe this correlation might be one of the causes of LBS outstanding performance. Further insights are provided in Appendix \ref{app:unsup-rew-perf}.

\begin{figure}[t!]
    \centering
    \begin{tabular}[t]{|l|c|c|c|c|}
    \hline
    \multicolumn{5}{|c|}{Pre-training for 2M environment frames}\\
    \hline
                                 & ICM & LBS      & P2E & RND \\ \hline
    Correlation & -0.54    & -0.60         & -0.34             & -0.03     \\ \hline
    p-value             & 0.07     & \textbf{0.04} & 0.28              & 0.91     \\ \hline
    \end{tabular}
    \captionof{table}{Pearson correlation and p-value between fine-tuned performance across URLB tasks and intrinsic rewards.}
    \label{tab:novelty}
\end{figure}

\section{Related Work}
\label{sec:citations}

\paragraph{Model-based control.}
Dynamics models combined with powerful search methods have led to impressive results on a wide variety of tasks such as Atari~\citep{schrittwieser2020mastering} and continuous control~\citep{Hafner2019Dream, Janner2019MBPO, sikchi2021learning, Lowrey2018POLO}. 
LOOP \citep{Silkchi2020LOOP} and TD-MPC \citep{Hansen2022TD-MPC} combine temporal difference learning and MPC. The model proposed with TD-MPC is task-oriented and thus requires a task to accelerate learning. In our work, we focus on unsupervised model learning, grounding on the DreamerV2 model \citep{Hafner2021DreamerV2}, whose supervision comes from predicting the environment's observations. Methods that use no reconstruction could generalize better to visual differences \citep{DreamerPro2021Deng, Ma2020CVRL} but they lose in explainability, for the absence of a decoder. 

\textbf{Unsupervised RL.}
Prior to our work, the large-scale study of curiosity \citep{Burda2018LSCuriosity} provided an insightful analysis of the performance of knowledge-based methods in the reward-free setting. In our work, we leverage the URLB setting, to provide an analysis of a combination of model-based control techniques with unsupervised RL. This allowed us to formulate a strategy to adapt pre-trained models to visual control tasks in a data-efficient manner. 
Closely, Plan2Explore \citep{sekar2020planning} adapts Disagreement \citep{pathak19a_disagreement} to work with Dreamer~\citep{Hafner2019Dream}. In our work, in addition to analyzing a wider choice of unsupervised RL strategies that can work with a world-model-based agent, we show how to better exploit the agent PT components for adaptation, and we propose a hybrid planner to improve data efficiency. As a results, we largely improved performance compared to Plan2Explore.

\textbf{Transfer learning.} In the field of transfer learning, fine-tuning is the most used approach. However, fine-tuning all the pre-trained agent components may not be the most effective strategy. In transfer learning for RL, they have studied this problem, mainly with the objective of transferring from one environment to another \citep{Farebrother2018DQNGener,Sasso2022MSTransfer, vanDriessel2021ComponentTransfer}. Instead, we analyze which agent's components should be transferred from the unsupervised PT stage to the supervised FT stage when the environment's dynamics is assumed to stay similar or be the same. 

\section{Conclusion}
\label{sec:conclusion}

In order to accelerate the development and deployment of learning agents for real-world tasks, it is crucial that the employed algorithms can adapt in a data-efficient way for multiple tasks. Our method obtains near-optimal performance in URLB from pixels, which is a challenging benchmark that has been widely adopted in the community, and showed robustness to perturbations in the environment, on the RWRL benchmark. We also analyzed several aspects of the learned models, to understand what could be improved further in the future to ease the adaptation process. 

Our results, establishing a new state-of-the-art in URLB from pixels, could become a reference to push advances in URL further. At the same time, as we close the gap with supervised baselines, we show the need for new benchmarks in the community for further developing URL research. One main limitation of URLB is that the environment is almost identical between PT and FT and, as we have shown, this makes it easier for model-based approaches to learn how the environment works. Although our evaluation on the RWRL benchmark shows that unsupervised pre-training is still beneficial despite visual differences and perturbations, additional precautions are required to deploy these systems in the real world. To support this line of research, future benchmarks should include variations (both visual and in the dynamics) between PT and FT, testing the generalization capabilities of the agent's behavior or the world model.

Another takeaway from our study on the URLB is the importance of learning generalizable and adaptable behaviors. In RL, this is reflected in the actor and critic models. While we showed that a critic model trained on unsupervised rewards cannot easily transfer to new tasks, there are lines of research that aim to learn generalizable critic networks by adopting successor features \citep{Hansen2019VISR,Barreto2016SFTransfer} or goal-directed value functions \citep{Ma2022VIP}, which could lead to faster or zero-shot adaptation to a given reward function \cite{Touati2023ZeroShot}. As for the actor, we believe that learning generalizable skills could be one potential way to learn more varied and adaptable action behaviors \citep{Pertsch2020SPIRL}. However, we also experienced that skill-driven methods \citep{aps, eysenbach2018diversity} tend to have more limited exploration capabilities \citep{Campos2020EDL} falling behind other approaches in some domains (i.e. Walker and Quadruped) of URLB.






\section*{Acknowledgements}
This research received funding from the Flemish Government (AI Research Program). Pietro Mazzaglia is funded by a Ph.D. grant of the Flanders Research Foundation (FWO). The authors would like to thank Sebastien Paquet and Chris Pal for their valuable feedback.



\bibliography{example_paper}

\begin{thebibliography}{59}
\providecommand{\natexlab}[1]{#1}
\providecommand{\url}[1]{\texttt{#1}}
\expandafter\ifx\csname urlstyle\endcsname\relax
  \providecommand{\doi}[1]{doi: #1}\else
  \providecommand{\doi}{doi: \begingroup \urlstyle{rm}\Url}\fi

\bibitem[Ahn et~al.(2022)Ahn, Brohan, Brown, Chebotar, Cortes, David, Finn,
  Gopalakrishnan, Hausman, Herzog, Ho, Hsu, Ibarz, Ichter, Irpan, Jang, Ruano,
  Jeffrey, Jesmonth, Joshi, Julian, Kalashnikov, Kuang, Lee, Levine, Lu, Luu,
  Parada, Pastor, Quiambao, Rao, Rettinghouse, Reyes, Sermanet, Sievers, Tan,
  Toshev, Vanhoucke, Xia, Xiao, Xu, Xu, and Yan]{Ahn2022DoAsICan}
Ahn, M., Brohan, A., Brown, N., Chebotar, Y., Cortes, O., David, B., Finn, C.,
  Gopalakrishnan, K., Hausman, K., Herzog, A., Ho, D., Hsu, J., Ibarz, J.,
  Ichter, B., Irpan, A., Jang, E., Ruano, R.~J., Jeffrey, K., Jesmonth, S.,
  Joshi, N., Julian, R., Kalashnikov, D., Kuang, Y., Lee, K.-H., Levine, S.,
  Lu, Y., Luu, L., Parada, C., Pastor, P., Quiambao, J., Rao, K., Rettinghouse,
  J., Reyes, D., Sermanet, P., Sievers, N., Tan, C., Toshev, A., Vanhoucke, V.,
  Xia, F., Xiao, T., Xu, P., Xu, S., and Yan, M.
\newblock Do as i can and not as i say: Grounding language in robotic
  affordances.
\newblock In \emph{arXiv preprint arXiv:2204.01691}, 2022.

\bibitem[Argenson \& Dulac-Arnold(2020)Argenson and
  Dulac-Arnold]{Argenson2020MOPO}
Argenson, A. and Dulac-Arnold, G.
\newblock Model-based offline planning, 2020.
\newblock URL \url{https://arxiv.org/abs/2008.05556}.

\bibitem[Barreto et~al.(2016)Barreto, Dabney, Munos, Hunt, Schaul, van Hasselt,
  and Silver]{Barreto2016SFTransfer}
Barreto, A., Dabney, W., Munos, R., Hunt, J.~J., Schaul, T., van Hasselt, H.,
  and Silver, D.
\newblock Successor features for transfer in reinforcement learning, 2016.
\newblock URL \url{https://arxiv.org/abs/1606.05312}.

\bibitem[Bellemare et~al.(2016)Bellemare, Srinivasan, Ostrovski, Schaul,
  Saxton, and Munos]{bellemarecount}
Bellemare, M., Srinivasan, S., Ostrovski, G., Schaul, T., Saxton, D., and
  Munos, R.
\newblock Unifying count-based exploration and intrinsic motivation.
\newblock In Lee, D., Sugiyama, M., Luxburg, U., Guyon, I., and Garnett, R.
  (eds.), \emph{Advances in Neural Information Processing Systems}, volume~29,
  2016.

\bibitem[Burda et~al.(2018)Burda, Edwards, Pathak, Storkey, Darrell, and
  Efros]{Burda2018LSCuriosity}
Burda, Y., Edwards, H., Pathak, D., Storkey, A., Darrell, T., and Efros, A.~A.
\newblock Large-scale study of curiosity-driven learning, 2018.
\newblock URL \url{https://arxiv.org/abs/1808.04355}.

\bibitem[Burda et~al.(2019{\natexlab{a}})Burda, Edwards, Pathak, Storkey,
  Darrell, and Efros]{Burda19}
Burda, Y., Edwards, H., Pathak, D., Storkey, A., Darrell, T., and Efros, A.~A.
\newblock Largescale study of curiosity-driven learning.
\newblock \emph{ICLR}, 2019{\natexlab{a}}.

\bibitem[Burda et~al.(2019{\natexlab{b}})Burda, Edwards, Storkey, and
  Klimov]{rnd}
Burda, Y., Edwards, H., Storkey, A.~J., and Klimov, O.
\newblock Exploration by random network distillation.
\newblock \emph{ICLR}, 2019{\natexlab{b}}.

\bibitem[Campos et~al.(2020)Campos, Trott, Xiong, Socher, Giro-i Nieto, and
  Torres]{Campos2020EDL}
Campos, V., Trott, A., Xiong, C., Socher, R., Giro-i Nieto, X., and Torres, J.
\newblock Explore, discover and learn: Unsupervised discovery of state-covering
  skills, 2020.
\newblock URL \url{https://arxiv.org/abs/2002.03647}.

\bibitem[Chen et~al.(2020)Chen, Kornblith, Norouzi, and Hinton]{Chen2020SimCLR}
Chen, T., Kornblith, S., Norouzi, M., and Hinton, G.
\newblock A simple framework for contrastive learning of visual
  representations.
\newblock In \emph{Proceedings of the 37th International Conference on Machine
  Learning}, pp.\  1597--1607, 2020.

\bibitem[Chua et~al.(2018)Chua, Calandra, McAllister, and
  Levine]{Chua2018PETS_MPC}
Chua, K., Calandra, R., McAllister, R., and Levine, S.
\newblock Deep reinforcement learning in a handful of trials using
  probabilistic dynamics models.
\newblock In Bengio, S., Wallach, H., Larochelle, H., Grauman, K.,
  Cesa-Bianchi, N., and Garnett, R. (eds.), \emph{Advances in Neural
  Information Processing Systems}. Curran Associates, Inc., 2018.

\bibitem[Chung et~al.(2014)Chung, Gulcehre, Cho, and Bengio]{gru14}
Chung, J., Gulcehre, C., Cho, K., and Bengio, Y.
\newblock Empirical evaluation of gated recurrent neural networks on sequence
  modeling.
\newblock In \emph{NIPS Workshop on Deep Learning, 2014}, 2014.

\bibitem[Deng et~al.(2021)Deng, Jang, and Ahn]{DreamerPro2021Deng}
Deng, F., Jang, I., and Ahn, S.
\newblock Dreamerpro: Reconstruction-free model-based reinforcement learning
  with prototypical representations, 2021.
\newblock URL \url{https://arxiv.org/abs/2110.14565}.

\bibitem[Dulac-Arnold et~al.(2020)Dulac-Arnold, Levine, Mankowitz, Li,
  Paduraru, Gowal, and Hester]{DulacArnold2020RWRL}
Dulac-Arnold, G., Levine, N., Mankowitz, D.~J., Li, J., Paduraru, C., Gowal,
  S., and Hester, T.
\newblock An empirical investigation of the challenges of real-world
  reinforcement learning, 2020.

\bibitem[Eysenbach et~al.(2019)Eysenbach, Gupta, Ibarz, and
  Levine]{eysenbach2018diversity}
Eysenbach, B., Gupta, A., Ibarz, J., and Levine, S.
\newblock Diversity is all you need: Learning skills without a reward function.
\newblock In \emph{ICLR}, 2019.

\bibitem[Farebrother et~al.(2018)Farebrother, Machado, and
  Bowling]{Farebrother2018DQNGener}
Farebrother, J., Machado, M.~C., and Bowling, M.
\newblock Generalization and regularization in dqn, 2018.
\newblock URL \url{https://arxiv.org/abs/1810.00123}.

\bibitem[Ha \& Schmidhuber(2018)Ha and Schmidhuber]{worldmodel}
Ha, D. and Schmidhuber, J.
\newblock Recurrent world models facilitate policy evolution.
\newblock In Bengio, S., Wallach, H., Larochelle, H., Grauman, K.,
  Cesa-Bianchi, N., and Garnett, R. (eds.), \emph{Advances in Neural
  Information Processing Systems}. Curran Associates, Inc., 2018.

\bibitem[Haarnoja et~al.(2018)Haarnoja, Zhou, Abbeel, and
  Levine]{haarnoja2018soft}
Haarnoja, T., Zhou, A., Abbeel, P., and Levine, S.
\newblock Soft actor-critic: Off-policy maximum entropy deep reinforcement
  learning with a stochastic actor.
\newblock \emph{arXiv preprint arXiv:1801.01290}, 2018.

\bibitem[Hafner et~al.(2019{\natexlab{a}})Hafner, Lillicrap, Ba, and
  Norouzi]{Hafner2019Dream}
Hafner, D., Lillicrap, T., Ba, J., and Norouzi, M.
\newblock Dream to control: Learning behaviors by latent imagination.
\newblock In \emph{ICLR}, 2019{\natexlab{a}}.

\bibitem[Hafner et~al.(2019{\natexlab{b}})Hafner, Lillicrap, Fischer, Villegas,
  Ha, Lee, and Davidson]{hafner2019planet}
Hafner, D., Lillicrap, T., Fischer, I., Villegas, R., Ha, D., Lee, H., and
  Davidson, J.
\newblock Learning latent dynamics for planning from pixels.
\newblock In \emph{ICML}, pp.\  2555--2565, 2019{\natexlab{b}}.

\bibitem[Hafner et~al.(2021)Hafner, Lillicrap, Norouzi, and
  Ba]{Hafner2021DreamerV2}
Hafner, D., Lillicrap, T.~P., Norouzi, M., and Ba, J.
\newblock Mastering atari with discrete world models.
\newblock In \emph{ICLR}, 2021.

\bibitem[Hansen et~al.(2022)Hansen, Wang, and Su]{Hansen2022TD-MPC}
Hansen, N., Wang, X., and Su, H.
\newblock Temporal difference learning for model predictive control.
\newblock 2022.

\bibitem[Hansen et~al.(2020)Hansen, Dabney, Barreto, Warde-Farley, de~Wiele,
  and Mnih]{Hansen2019VISR}
Hansen, S., Dabney, W., Barreto, A., Warde-Farley, D., de~Wiele, T.~V., and
  Mnih, V.
\newblock Fast task inference with variational intrinsic successor features.
\newblock In \emph{ICLR}, 2020.

\bibitem[Janner et~al.(2019)Janner, Fu, Zhang, and Levine]{Janner2019MBPO}
Janner, M., Fu, J., Zhang, M., and Levine, S.
\newblock When to trust your model: Model-based policy optimization.
\newblock \emph{ArXiv}, abs/1906.08253, 2019.

\bibitem[Laskin et~al.(2021)Laskin, Yarats, Liu, Lee, Zhan, Lu, Cang, Pinto,
  and Abbeel]{laskin2021urlb}
Laskin, M., Yarats, D., Liu, H., Lee, K., Zhan, A., Lu, K., Cang, C., Pinto,
  L., and Abbeel, P.
\newblock {URLB}: Unsupervised reinforcement learning benchmark.
\newblock In \emph{NeurIPS Datasets and Benchmarks Track (Round 2)}, 2021.

\bibitem[LeCun(2022)]{lecun2022path}
LeCun, Y.
\newblock A path towards autonomous machine intelligence version 0.9. 2,
  2022-06-27.
\newblock \emph{Open Review}, 62, 2022.

\bibitem[Levine et~al.(2016)Levine, Finn, Darrell, and Abbeel]{levine2016}
Levine, S., Finn, C., Darrell, T., and Abbeel, P.
\newblock End-to-end training of deep visuomotor policies.
\newblock \emph{J. Mach. Learn. Res.}, 2016.

\bibitem[Lillicrap et~al.(2016)Lillicrap, Hunt, Pritzel, Heess, Erez, Tassa,
  Silver, and Wierstra]{LillicrapHPHETS15}
Lillicrap, T.~P., Hunt, J.~J., Pritzel, A., Heess, N., Erez, T., Tassa, Y.,
  Silver, D., and Wierstra, D.
\newblock Continuous control with deep reinforcement learning.
\newblock In Bengio, Y. and LeCun, Y. (eds.), \emph{ICLR}, 2016.

\bibitem[Liu \& Abbeel(2021{\natexlab{a}})Liu and Abbeel]{aps}
Liu, H. and Abbeel, P.
\newblock Aps: Active pretraining with successor features.
\newblock In \emph{ICML}, pp.\  6736--6747, 2021{\natexlab{a}}.

\bibitem[Liu \& Abbeel(2021{\natexlab{b}})Liu and
  Abbeel]{liu2021unsupervised_apt}
Liu, H. and Abbeel, P.
\newblock Unsupervised active pre-training for reinforcement learning.
\newblock \emph{ICLR}, 2021{\natexlab{b}}.

\bibitem[Lowrey et~al.(2018)Lowrey, Rajeswaran, Kakade, Todorov, and
  Mordatch]{Lowrey2018POLO}
Lowrey, K., Rajeswaran, A., Kakade, S., Todorov, E., and Mordatch, I.
\newblock Plan online, learn offline: Efficient learning and exploration via
  model-based control, 2018.
\newblock URL \url{https://arxiv.org/abs/1811.01848}.

\bibitem[Lu et~al.(2021)Lu, Hausman, Chebotar, Yan, Jang, Herzog, Xiao, Irpan,
  Khansari, Kalashnikov, and Levine]{Lu2021AWOpt}
Lu, Y., Hausman, K., Chebotar, Y., Yan, M., Jang, E., Herzog, A., Xiao, T.,
  Irpan, A., Khansari, M., Kalashnikov, D., and Levine, S.
\newblock {AW}-opt: Learning robotic skills with imitation andreinforcement at
  scale.
\newblock In \emph{5th Annual Conference on Robot Learning (CoRL)}, 2021.

\bibitem[Ma et~al.(2020)Ma, Chen, Hsu, and Lee]{Ma2020CVRL}
Ma, X., Chen, S., Hsu, D., and Lee, W.~S.
\newblock Contrastive variational model-based reinforcement learning for
  complex observations.
\newblock In \emph{Proceedings of the 4th Conference on Robot Learning (CoRL)},
  2020.

\bibitem[Ma et~al.(2022)Ma, Sodhani, Jayaraman, Bastani, Kumar, and
  Zhang]{Ma2022VIP}
Ma, Y.~J., Sodhani, S., Jayaraman, D., Bastani, O., Kumar, V., and Zhang, A.
\newblock Vip: Towards universal visual reward and representation via
  value-implicit pre-training, 2022.
\newblock URL \url{https://arxiv.org/abs/2210.00030}.

\bibitem[Mazzaglia et~al.(2021)Mazzaglia, Çatal, Verbelen, and
  Dhoedt]{Mazzaglia2021SelfSupervisedEV}
Mazzaglia, P., Çatal, O., Verbelen, T., and Dhoedt, B.
\newblock Curiosity-driven exploration via latent bayesian surprise.
\newblock \emph{ArXiv}, abs/2104.07495, 2021.

\bibitem[OpenAI et~al.(2019)OpenAI, Akkaya, Andrychowicz, Chociej, Litwin,
  McGrew, Petron, Paino, Plappert, Powell, Ribas, Schneider, Tezak, Tworek,
  Welinder, Weng, Yuan, Zaremba, and Zhang]{OpenAI2019Rubik}
OpenAI, Akkaya, I., Andrychowicz, M., Chociej, M., Litwin, M., McGrew, B.,
  Petron, A., Paino, A., Plappert, M., Powell, G., Ribas, R., Schneider, J.,
  Tezak, N.~A., Tworek, J., Welinder, P., Weng, L., Yuan, Q., Zaremba, W., and
  Zhang, L.~M.
\newblock Solving rubik's cube with a robot hand.
\newblock \emph{ArXiv}, abs/1910.07113, 2019.

\bibitem[Parisi et~al.(2022)Parisi, Rajeswaran, Purushwalkam, and
  Gupta]{Parisi2022PTVisionRobot}
Parisi, S., Rajeswaran, A., Purushwalkam, S., and Gupta, A.
\newblock The unsurprising effectiveness of pre-trained vision models for
  control, 2022.

\bibitem[Pathak et~al.(2017)Pathak, Agrawal, Efros, and
  Darrell]{Pathak_curiosity}
Pathak, D., Agrawal, P., Efros, A., and Darrell, T.
\newblock Curiosity-driven exploration by self-supervised prediction.
\newblock \emph{ICML}, 2017.

\bibitem[Pathak et~al.(2019)Pathak, Gandhi, and Gupta]{pathak19a_disagreement}
Pathak, D., Gandhi, D., and Gupta, A.
\newblock Self-supervised exploration via disagreement.
\newblock In \emph{ICML}, 2019.

\bibitem[Pertsch et~al.(2020)Pertsch, Lee, and Lim]{Pertsch2020SPIRL}
Pertsch, K., Lee, Y., and Lim, J.~J.
\newblock Accelerating reinforcement learning with learned skill priors, 2020.
\newblock URL \url{https://arxiv.org/abs/2010.11944}.

\bibitem[Radford et~al.(2019)Radford, Wu, Child, Luan, Amodei, and
  Sutskever]{Radford2019GPT2}
Radford, A., Wu, J., Child, R., Luan, D., Amodei, D., and Sutskever, I.
\newblock Language models are unsupervised multitask learners.
\newblock 2019.

\bibitem[Richards(2005)]{Richards2005RobustMPC}
Richards, A.~G.
\newblock \emph{Robust constrained model predictive control}.
\newblock PhD thesis, Massachusetts Institute of Technology, 2005.

\bibitem[Rubinstein \& Kroese(2004)Rubinstein and Kroese]{rubinstein2004cross}
Rubinstein, R.~Y. and Kroese, D.~P.
\newblock \emph{The cross-entropy method: a unified approach to combinatorial
  optimization, Monte-Carlo simulation, and machine learning}, volume 133.
\newblock Springer, 2004.

\bibitem[Sasso et~al.(2022)Sasso, Sabatelli, and Wiering]{Sasso2022MSTransfer}
Sasso, R., Sabatelli, M., and Wiering, M.~A.
\newblock Multi-source transfer learning for deep model-based reinforcement
  learning, 2022.
\newblock URL \url{https://arxiv.org/abs/2205.14410}.

\bibitem[Schrittwieser et~al.(2020)Schrittwieser, Antonoglou, Hubert, Simonyan,
  Sifre, Schmitt, Guez, Lockhart, Hassabis, Graepel,
  et~al.]{schrittwieser2020mastering}
Schrittwieser, J., Antonoglou, I., Hubert, T., Simonyan, K., Sifre, L.,
  Schmitt, S., Guez, A., Lockhart, E., Hassabis, D., Graepel, T., et~al.
\newblock Mastering atari, go, chess and shogi by planning with a learned
  model.
\newblock \emph{Nature}, 588\penalty0 (7839):\penalty0 604--609, 2020.

\bibitem[Schulman et~al.(2016)Schulman, Moritz, Levine, Jordan, and
  Abbeel]{Schulman2015GAE}
Schulman, J., Moritz, P., Levine, S., Jordan, M., and Abbeel, P.
\newblock High-dimensional continuous control using generalized advantage
  estimation.
\newblock In \emph{ICLR}, 2016.

\bibitem[Schulman et~al.(2017)Schulman, Wolski, Dhariwal, Radford, and
  Klimov]{Schulman2017PPO}
Schulman, J., Wolski, F., Dhariwal, P., Radford, A., and Klimov, O.
\newblock Proximal policy optimization algorithms.
\newblock \emph{ArXiv}, abs/1707.06347, 2017.

\bibitem[Sekar et~al.(2020)Sekar, Rybkin, Daniilidis, Abbeel, Hafner, and
  Pathak]{sekar2020planning}
Sekar, R., Rybkin, O., Daniilidis, K., Abbeel, P., Hafner, D., and Pathak, D.
\newblock Planning to explore via self-supervised world models.
\newblock In \emph{ICML}, 2020.

\bibitem[Sikchi et~al.(2020)Sikchi, Zhou, and Held]{Silkchi2020LOOP}
Sikchi, H., Zhou, W., and Held, D.
\newblock Learning off-policy with online planning, 2020.
\newblock URL \url{https://arxiv.org/abs/2008.10066}.

\bibitem[Sikchi et~al.(2021)Sikchi, Zhou, and Held]{sikchi2021learning}
Sikchi, H., Zhou, W., and Held, D.
\newblock Learning off-policy with online planning.
\newblock In \emph{5th Annual Conference on Robot Learning}, 2021.
\newblock URL \url{https://openreview.net/forum?id=1GNV9SW95eJ}.

\bibitem[Singh et~al.(2003)Singh, Hnizdo, Demchuk, and Misra]{Singh2003PBE}
Singh, H., Hnizdo, V., Demchuk, A., and Misra, N.
\newblock Nearest neighbor estimates of entropy.
\newblock \emph{American Journal of Mathematical and Management Sciences}, 23,
  02 2003.

\bibitem[Sutton(1991)]{Sutton1991Dyna}
Sutton, R.~S.
\newblock Dyna, an integrated architecture for learning, planning, and
  reacting.
\newblock \emph{SIGART Bull.}, 2\penalty0 (4):\penalty0 160–163, jul 1991.
\newblock ISSN 0163-5719.
\newblock \doi{10.1145/122344.122377}.
\newblock URL \url{https://doi.org/10.1145/122344.122377}.

\bibitem[Talvitie(2018)]{Talvitie2018RewardMiss}
Talvitie, E.
\newblock Learning the reward function for a misspecified model.
\newblock In Dy, J. and Krause, A. (eds.), \emph{ICML}, volume~80 of
  \emph{Proceedings of Machine Learning Research}, pp.\  4838--4847. PMLR,
  10--15 Jul 2018.

\bibitem[Tassa et~al.(2018)Tassa, Doron, Muldal, Erez, Li, de~Las~Casas,
  Budden, Abdolmaleki, Merel, Lefrancq, Lillicrap, and
  Riedmiller]{tassa_dmcontrol}
Tassa, Y., Doron, Y., Muldal, A., Erez, T., Li, Y., de~Las~Casas, D., Budden,
  D., Abdolmaleki, A., Merel, J., Lefrancq, A., Lillicrap, T.~P., and
  Riedmiller, M.~A.
\newblock Deepmind control suite.
\newblock \emph{CoRR}, abs/1801.00690, 2018.

\bibitem[Touati et~al.(2023)Touati, Rapin, and Ollivier]{Touati2023ZeroShot}
Touati, A., Rapin, J., and Ollivier, Y.
\newblock Does zero-shot reinforcement learning exist?, 2023.

\bibitem[van Driessel \& Francois-Lavet(2021)van Driessel and
  Francois-Lavet]{vanDriessel2021ComponentTransfer}
van Driessel, G. and Francois-Lavet, V.
\newblock Component transfer learning for deep rl based on abstract
  representations, 2021.
\newblock URL \url{https://arxiv.org/abs/2111.11525}.

\bibitem[{Williams} et~al.(2015){Williams}, {Aldrich}, and
  {Theodorou}]{Williams2015MPPI}
{Williams}, G., {Aldrich}, A., and {Theodorou}, E.
\newblock {Model Predictive Path Integral Control using Covariance Variable
  Importance Sampling}.
\newblock \emph{arXiv e-prints}, art. arXiv:1509.01149, 2015.

\bibitem[Wu et~al.(2022)Wu, Escontrela, Hafner, Goldberg, and
  Abbeel]{Wu2022DayDreamer}
Wu, P., Escontrela, A., Hafner, D., Goldberg, K., and Abbeel, P.
\newblock Daydreamer: World models for physical robot learning, 2022.
\newblock URL \url{https://arxiv.org/abs/2206.14176}.

\bibitem[Yarats et~al.(2021)Yarats, Fergus, Lazaric, and
  Pinto]{yarats2021proto}
Yarats, D., Fergus, R., Lazaric, A., and Pinto, L.
\newblock Reinforcement learning with prototypical representations.
\newblock 2021.

\bibitem[Yarats et~al.(2022)Yarats, Fergus, Lazaric, and
  Pinto]{Yarats2021DrQ-v2}
Yarats, D., Fergus, R., Lazaric, A., and Pinto, L.
\newblock Mastering visual continuous control: Improved data-augmented
  reinforcement learning.
\newblock In \emph{ICLR}, 2022.

\end{thebibliography}
\bibliographystyle{icml2023}

\clearpage
\newpage

\onecolumn

\appendix
\section*{Appendix}
\section{Normalization scores}
\label{app:reference_scores}



\begin{table*}[h]
\centering
\resizebox{\linewidth}{!}{
\begin{tabular}{|lc|cc|cc|}
\hline
\multicolumn{6}{|c|}{Pre-trainining for 2M environment frames}\\
\hline
Domain & Task & URLB Expert & URLB Disagreement &  Dreamer@2M &  Ours \\
\hline
\multirow{4}{*}{Walker}&  Flip & 799 &  339 $\pm$ 16\phantom{0} & 778 & 938 $\pm$ 5 \\
 &  Run & 796& 154 $\pm$ 9\phantom{0} & 724 & 596 $\pm$ 17 \\
 &  Stand & 984 & 552 $\pm$ 92 &  909 & 973 $ \pm $ 6\phantom{0} \\
 &  Walk & 971& 424 $\pm$ 36 &  965  & 959 $\pm$ 0\phantom{0} \\
\hline
\multirow{4}{*}{Quadruped}&  Jump & 888 & 194 $\pm$ 21 & 753 & 822 $\pm$ 15 \\
 &  Run & 888 & 143 $\pm$ 25 &  904 & 642 $\pm$ 44  \\
 &  Stand & 920& 305 $\pm$ 35 & 945 & 927 $\pm$ 13   \\
 &  Walk & 866 & 145 $\pm$ 10 &  947 & 816 $\pm$ 27   \\
\hline
\multirow{4}{*}{Jaco}&  Reach bottom left & 193 & 106 $\pm$ 22 &  223 & 192 $\pm$ 10 \\
 &  Reach bottom right & 203 & 90 $\pm$ 15 &  231 & 192 $\pm$ 8\phantom{0} \\
 &  Reach top left & 191 & 127 $\pm$ 21 &  233  & 197 $\pm$ 8\phantom{0} \\
 &  Reach top right & 223 & 118 $\pm$ 23  &  225 & 212 $\pm$ 6\phantom{0} \\
\hline
\end{tabular}}
\vspace{1em}
\caption{Performance of expert baseline and the best method on pixel-based URLB from \cite{laskin2021urlb} and performance of our oracle baseline (Dreamer@2M) and best approach, using LBS for unsupervised data collection, after pre-training for 2M frames and fine-tuning for 100k steps.}
\label{table:states_numbers}
\end{table*}
 

In Table \ref{table:states_numbers}, we report the mean scores for the URLB Expert, used to normalize the scores in the URLB paper, and for Dreamer@2M, which we use to normalize returns of our methods, where both supervised baselines have been trained individually on each of the 12 tasks from URLB for 2M frames. We additionally report mean and standard errors for the best-performing unsupervised baseline from URLB. which is Disagreement~\citep{pathak19a_disagreement}, and our method. We notice that our scores approach the Dreamer@2M's scores in several tasks, eventually outperforming them in a few tasks (e.g. Walker Flip, Quadruped Jump). We believe this merit is due both to the exploration pre-training, which may have found more rewarding trajectories than greedy supervised RL optimization and of the improved Dyna-MPC planning strategy.

\section{Integrating Unsupervised RL Strategies}
\label{app:exploration}
We summarize here the unsupervised RL approaches tested and how we integrated them with the Dreamer algorithm for exploration. For all methods, rewards have been normalized during training using an exponential moving average with momentum $0.95$, with the exceptions of RND, which follows its original reward normalization \citep{rnd}, and APS, whose rewards are not normalized because they are used to regress the skill that is closer to the downstream task during FT.

\paragraph{ICM.} The Intrinsic Curiosity Module (ICM; \citet{Pathak_curiosity}) defines intrinsic rewards as the error between states projected in a feature space and a feature dynamics model's predictions. We use the Dreamer agent encoder $e_t = f_\phi(s_t)$ to obtain features and train a forward dynamics model $g(e_t| e_{t-1}, a_{t-1})$ to compute rewards as:
\begin{equation*}
    {r_t}^\text{ICM} \propto \Vert g(e_t| e_{t-1}, a_{t-1}) - e_t \Vert^2.
\end{equation*}
As the rewards for ICM require environment states (going through the encoder to compute prediction error), we train a reward predictor to allow estimating rewards in imagination. 

\paragraph{Plan2Explore.} The Plan2Explore algorithm~\citep{sekar2020planning} is an adaptation of the Disagreement algorithm~\citep{pathak19a_disagreement} for latent dynamics models. An ensemble of forward dynamics models is trained to predict the features embedding $e_t = f_\phi(s_t)$, given the previous latent state and actions, i.e. $g(e_t|z_{t-1}, a_{t-1}, w_k)$, where $w_k$ are the parameters of the k-th predictor. Intrinsic rewards are defined as the variance of the ensemble predictions:
\begin{equation*}
    {r_t}^\text{P2E} \propto \textrm{Var}(\{ g(e_t| z_{t-1}, a_{t-1}, w_k) | k \in [1, ..., K] \}).
\end{equation*}
Plan2Explore requires only latent states and actions, thus it can be computed directly in imagination. We used an ensemble of 5 models.

\paragraph{RND.}
Random Network Distillation (RND; \citet{rnd}) learns to predict the output of a randomly initialized network $n(s_t)$ that projects the states into a more compact random feature space. As the random network is not updated during training, the prediction error should diminish for already visited states. The intrinsic reward here is defined as: 
\begin{equation*}
    {r_t}^\text{RND} \propto \Vert g(s_t) - n(s_t) \Vert^2
\end{equation*}
As the rewards for RND requires environment states (to encode with the random network), we train a reward predictor to allow estimating rewards in imagination.

\paragraph{LBS.} In Latent Bayesian Surprise (LBS; \citet{Mazzaglia2021SelfSupervisedEV}), they use the KL divergence between the posterior and the prior of a latent dynamics model as a proxy for the information gained with respect to the latent state variable, by observing new states. Rewards are computed as:
\begin{equation*}
    {r_t}^\text{LBS} \propto \KL[q(z_t|z_{t-1}, a_{t-1}, e_t) \Vert p(z_t|z_{t-1}, a_{t-1})] 
\end{equation*}
As the rewards for LBS requires environment states (to compute the posterior distribution), we train a reward predictor to allow estimating rewards in imagination. 

\paragraph{APT.} Active Pre-training (APT; \citet{liu2021unsupervised_apt}) uses a particle-based estimator based on the K nearest-neighbors algorithm \citep{Singh2003PBE} to estimate entropy for a given state. We implement APT on top of the deterministic component of the latent states $\bar{z}_t$, providing rewards as:
\begin{equation*}
    {r_t}^\text{APT} \propto \sum_i^k \log \Vert \bar{z}_t - \bar{z}^i_t \Vert^2,
\end{equation*}
where $k$ are the nearest-neighbor states in latent space. As APT requires only latent states, it can be computed directly in imagination. We used $k=12$ nearest neighbors.

\paragraph{DIAYN.}
Diversity is All you need (DIAYN; \citet{eysenbach2018diversity}) maximizes the mutual information between the states and latent skills $w$. We implement DIAYN on top of the latent space of Dreamer, writing the mutual information as $I(w_t, z_t) = H(w_t) - H(w_t|z_t)$. The entropy $H(w_t)$ is kept maximal by sampling $w_t \sim \textrm{Unif}(w_t)$ from a discrete uniform prior distribution, while $H(w_t|z_t)$ is estimated learning a discriminator $q(w_t|z_t)$. We compute intrinsic rewards as:
\begin{equation*}
    {r_t}^\text{DIAYN} \propto \log q(w_t|z_t) 
\end{equation*}
Additionally, DIAYN maximizes the entropy of the actor, so we add an entropy maximization term to Dreamer's objective \citep{haarnoja2018soft}.
As DIAYN requires model states and skills sampled from a uniform distribution to compute rewards, we can directly compute them in imagination.
For FT, the skill adapted is the one with the highest expected rewards, considering the states and rewards obtained in the initial episodes.

\paragraph{APS.}
Active Pre-training with Successor features (APS; \citet{aps}) maximizes the mutual information between the states and latent skills $w$. We implement APS on top of the latent space of Dreamer, writing the mutual information as $I(w_t, z_t) = H(z_t) - H(z_t|w_t)$. The entropy term $H(z_t)$ is estimated using a particle-based estimator on top of the deterministic component of the latent states $\bar{z}_t$, as for APT, while the term $H(z_t|w_t)$ is estimated learning a discriminator $q(z_t|w_t)$. The intrinsic rewards for APS can be written as:
\begin{equation*}
    {r_t}^\text{APS} \propto {r_t}^\text{APT} + \log q(w_t|z_t)
\end{equation*}
As APS requires model states and uniformly sampled skills to compute rewards, we can directly compute them in imagination.
For FT, the skill to adapt is selected using linear regression over the states and rewards obtained in the initial episodes \citep{aps}.




\newpage

\section{Algorithm}
\label{app:algorithm}

\begin{algorithm*}[h]
\caption{\currentTitle
}\label{alg:urlb}
\begin{algorithmic}[1]
\Require Actor $\theta$, Critic $\psi$, World Model $\phi$ \\
~~~~~~~~~ Intrinsic reward $r^{\text{int}}$, extrinsic reward $r^{\text{ext}}$ \\
~~~~~~~~~ Environment, $M$, downstream tasks $T_k$, $k\in[1,\ldots,M]$ \\
~~~~~~~~~ Pre-train frames $N_{\text{PT}}$, fine-tune frames $N_{\text{FT}}$, environment frames/update $\tau$ \\
~~~~~~~~~ Initial model state $z_{0}$, hybrid planner $\text{Dyna-MPC}$, replay buffers $\gD_\text{PT}$, $\gD_\text{FT}$ \\
\\
\emph{// Pre-training}
\For{$t=0,\ldots,N_{\text{PT}}$}
    \State Draw action from the actor, $\rva_t \sim \pi_\theta(a_t|z_t)$
    \State Apply action to the environment, $\rvs_{t+1} \sim P(\cdot|\rvs_t, \rva_t)$
    \State Add transition to replay buffer, $\gD_\text{PT} \leftarrow \gD_\text{PT} \cup (\rvs_t, \rva_t, \rvs_{t+1})$
    \State Infer model state, $z_{t+1} \sim q(z_{t+1}|z_t, a_t, f_\phi(s_{t+1}))$
    \If{ $t \mod \tau = 0$ }
        \State Update world model parameters $\phi$ on the data from the replay buffer $\gD_\text{PT}$
        \State Update actor-critic parameters $\{\theta,\psi\}$ in imagination, maximizing $r^\text{int}$
    \EndIf
\EndFor
\State Output pre-trained parameters $\{\psi_\text{PT}, \theta_\text{PT},\phi_\text{PT}\}$ \\
\\
\emph{// Fine-tuning}
\For{$T_k \in [T_1, \ldots, T_M]$}
    \State Initialize fine-tuning world-model with $\phi_\text{PT}$
    \State (Optional) Initialize fine-tuning actor with $\theta_\text{PT}$
    \For{$t=0,\ldots,N_{\text{FT}}$}
        \State Draw action from the actor, $\rva_t \sim \pi_\theta(a_t|z_t)$
        \State Use the planner for selecting best action, $\rva_t \sim \text{Dyna-MPC}(z_t)$
        \State Apply action to the environment, $\rvs_{t+1}, r^{\text{ext}}_t \sim P(\cdot|\rvs_t, \rva_t)$
        \State Add transition to replay buffer, $\gD_\text{FT} \leftarrow \gD_\text{FT} \cup (\rvs_t, \rva_t, r^{\text{ext}}_t, \rvs_{t+1})$
        \State Infer model state, $z_{t+1} \sim q(z_{t+1}|z_t, a_t, f_\phi(s_{t+1}))$
        \If{ $t \mod \tau = 0$ }
            \State Update world model parameters $\phi$ on the data from the replay buffer $\gD_\text{FT}$
            \State Update actor-critic parameters $\{\theta,\psi\}$ in imagination, maximizing $r^\text{ext}$
        \EndIf
    \EndFor
    \State Evaluate performance on $T_k$
\EndFor
\end{algorithmic}
\end{algorithm*}

\clearpage
\newpage
\section{Additional Results}
\label{app:add_results}

We present complete results, for each unsupervised RL method, for the large-scale study experiments presented in Section \ref{sec:unsupervised_expl} and for the ablations in Section \ref{sec:result}.

\begin{figure*}[h!]
    \centering
    \includegraphics[width=\textwidth]{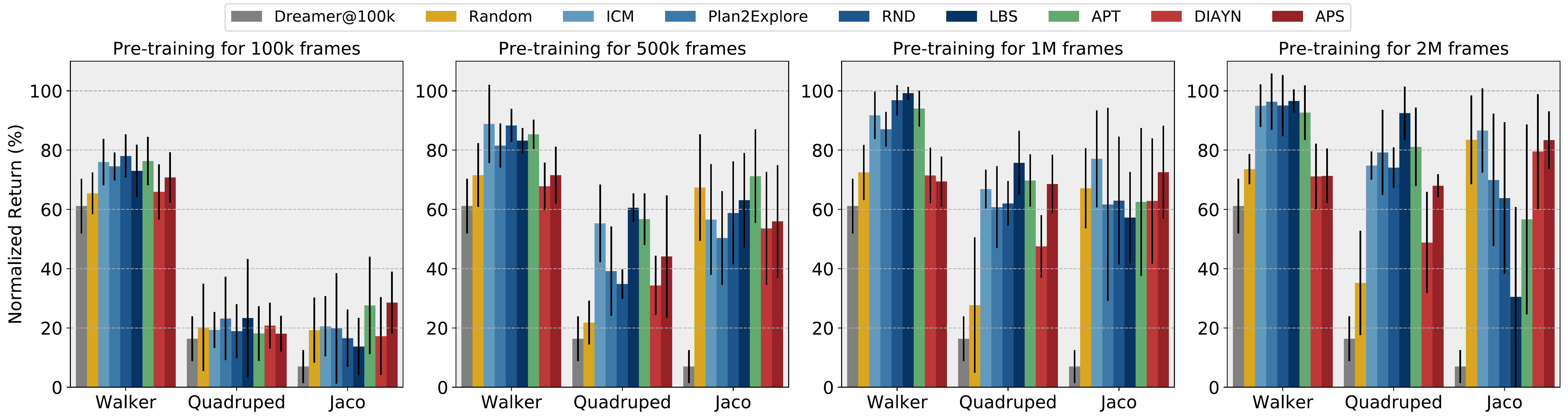}
    \caption{Complete results for Figure \ref{subfig:ours}.}
    \label{fig:expl_all_methods}
\end{figure*}

\begin{figure*}[h!]
\centering
\begin{subfigure}[t]{0.32\linewidth}
    \centering
    \includegraphics[width=\linewidth]{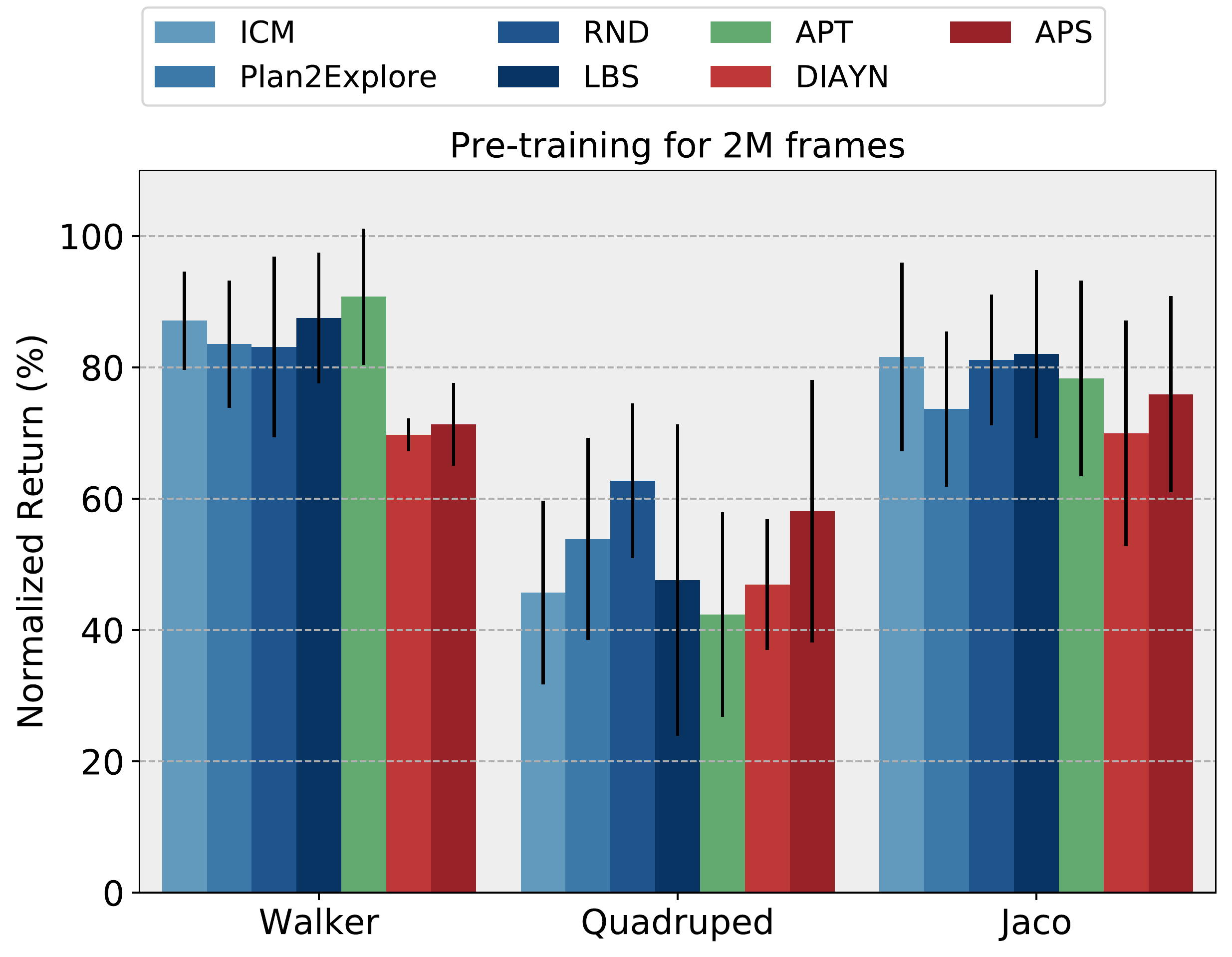}
    \caption{Model}
    \label{subfig:default}
\end{subfigure}
\begin{subfigure}[t]{0.32\linewidth}
    \centering
    \includegraphics[width=\linewidth]{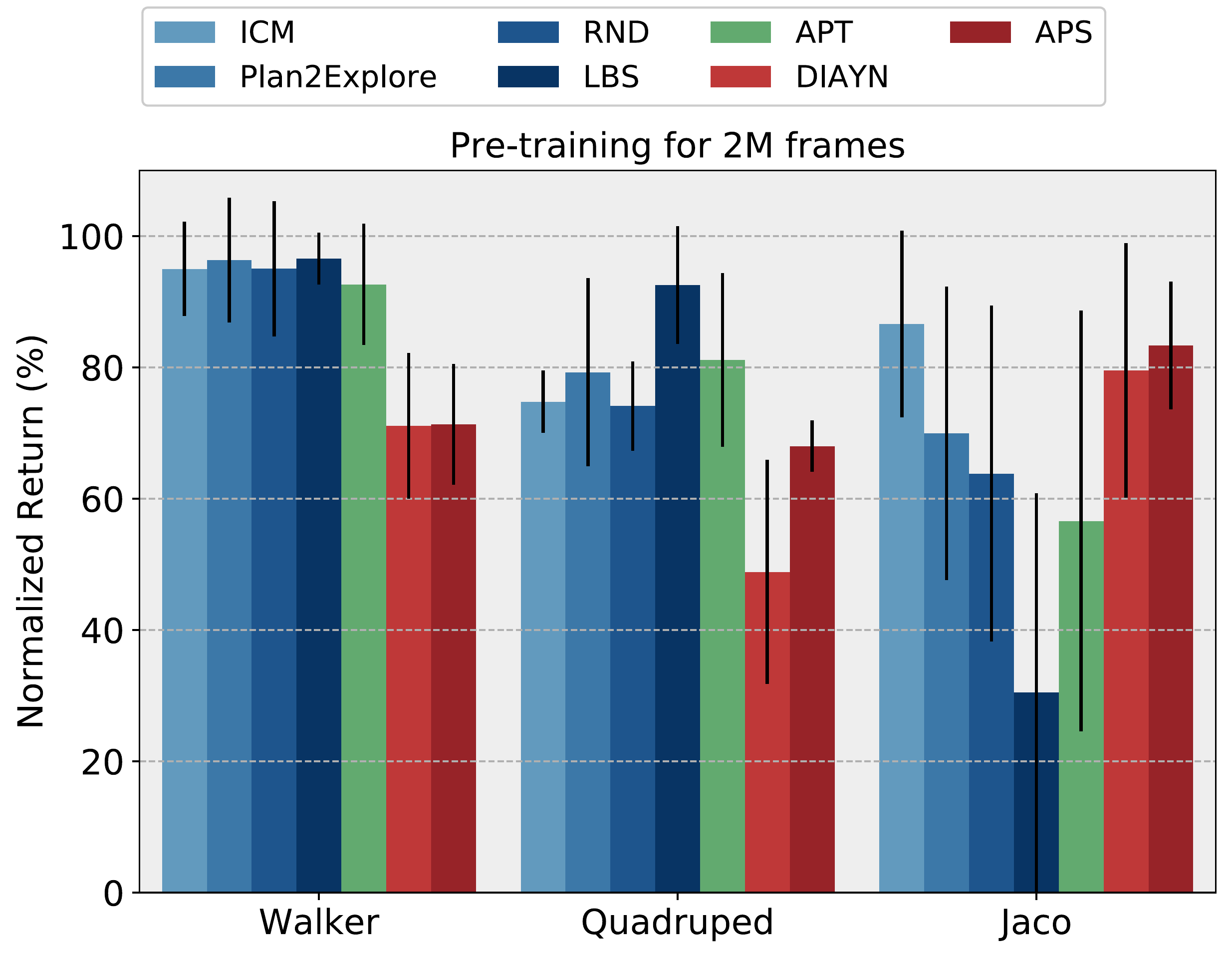}
    \caption{Model, Actor}
    \label{subfig:w_critic}
\end{subfigure}
\begin{subfigure}[t]{0.32\linewidth}
    \centering
    \includegraphics[width=\linewidth]{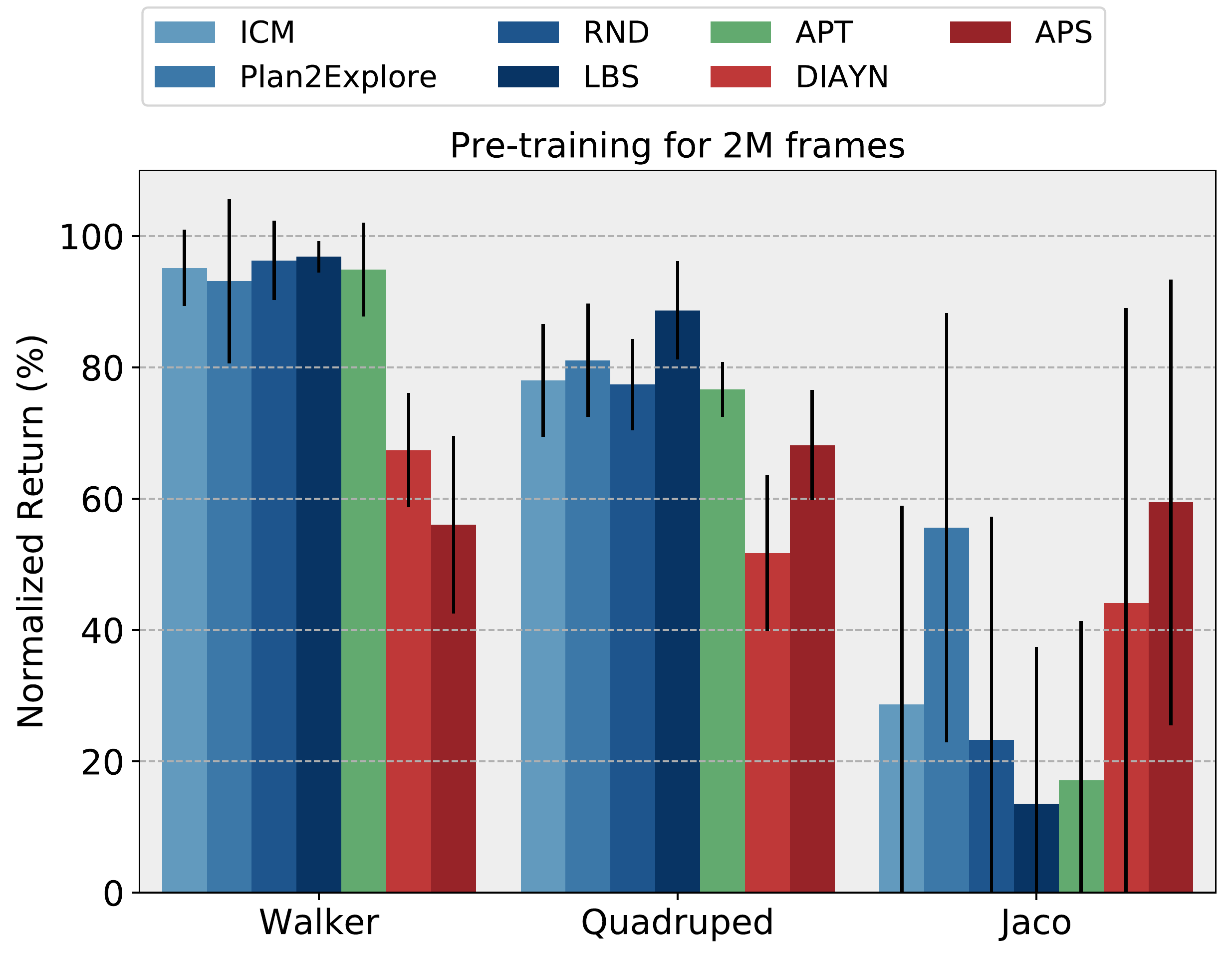}
    \caption{Model, Actor, Critic}
    \label{subfig:wo_policy}
\end{subfigure}
\caption{Complete results for Figure \ref{fig:modules_results}.}
\label{fig:complete_ac}
\end{figure*}

\begin{figure*}[h!]
\centering
\begin{subfigure}[t]{0.48\linewidth}
    \centering
    \includegraphics[width=\linewidth]{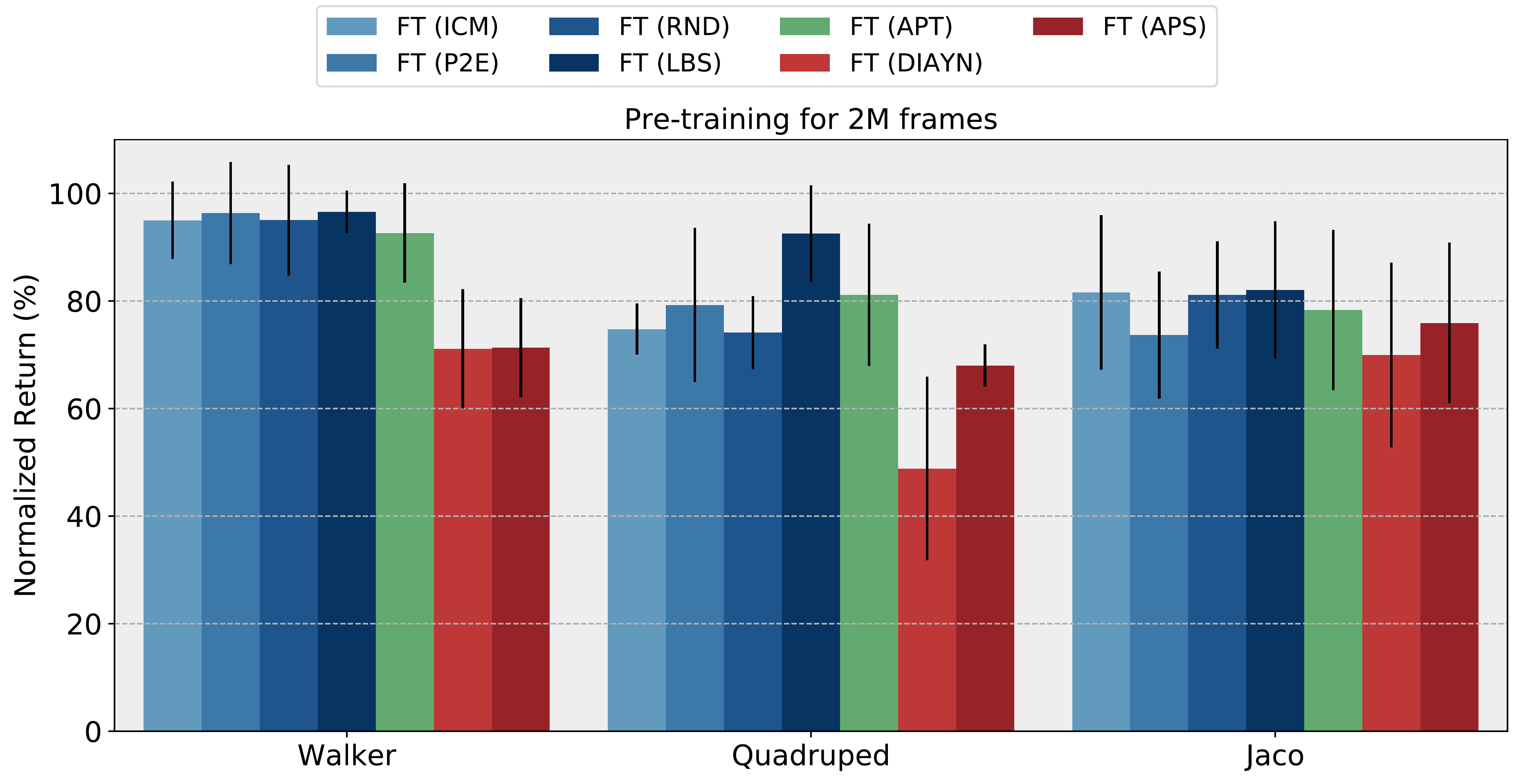}
    \caption{FT}
    \label{subfig:jaco_fix}
\end{subfigure}
\begin{subfigure}[t]{0.48\linewidth}
    \centering
    \includegraphics[width=\linewidth]{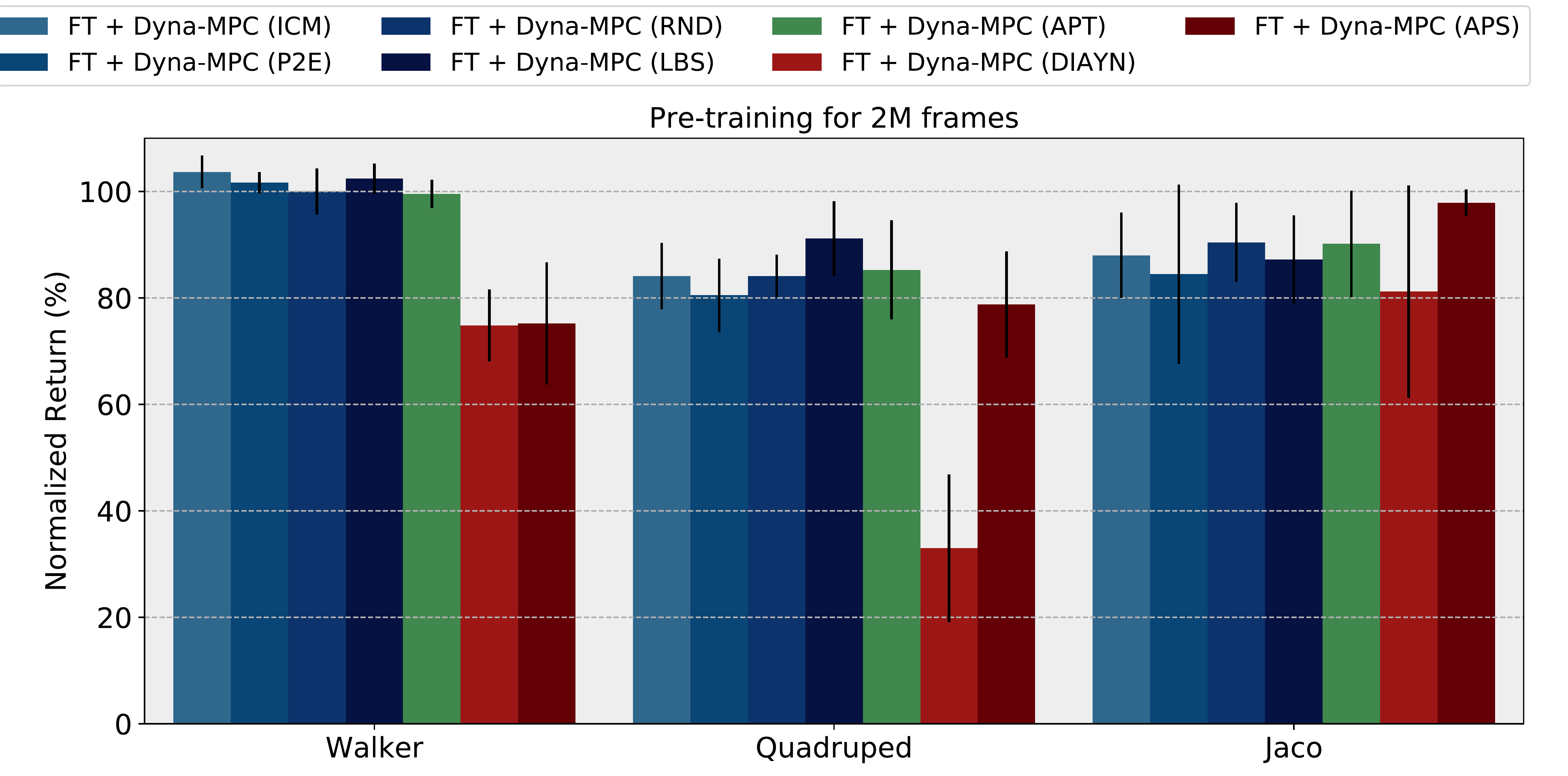}
    \caption{FT + Dyna-MPC}
    \label{subfig:dynampc_jaco_fix}
\end{subfigure}
\caption{Complete results for Figure \ref{fig:planning_improv}.}
\label{fig:complete_dynampc}
\end{figure*}

\newpage

\begin{figure*}[h]
    \centering
    \includegraphics[width=.9\linewidth]{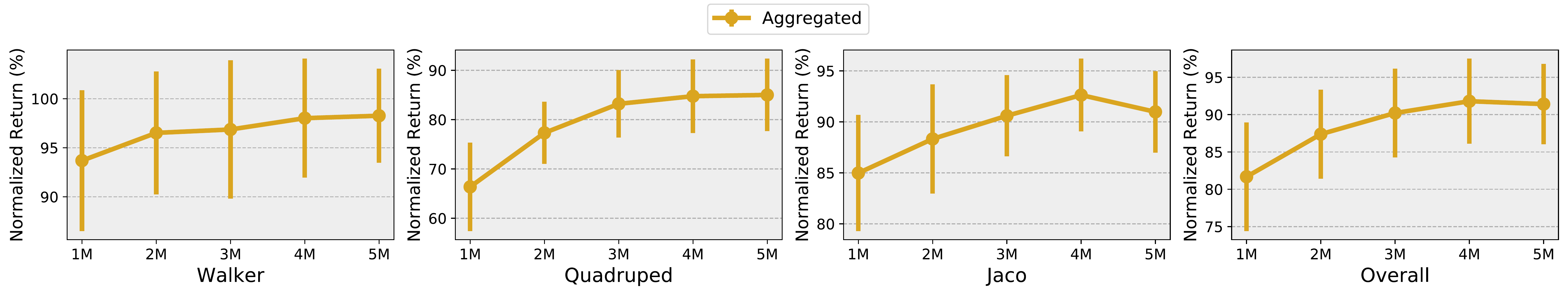}
    \includegraphics[width=.9\linewidth]{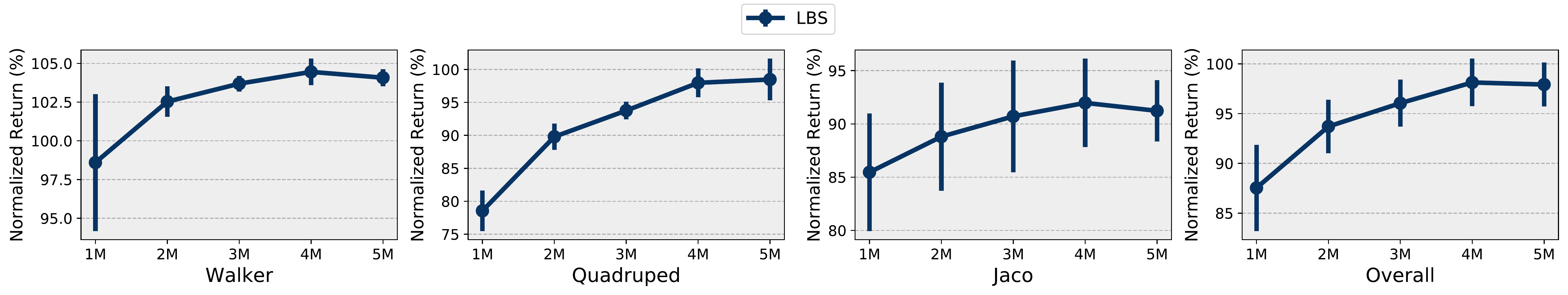}
    \includegraphics[width=.9\linewidth]{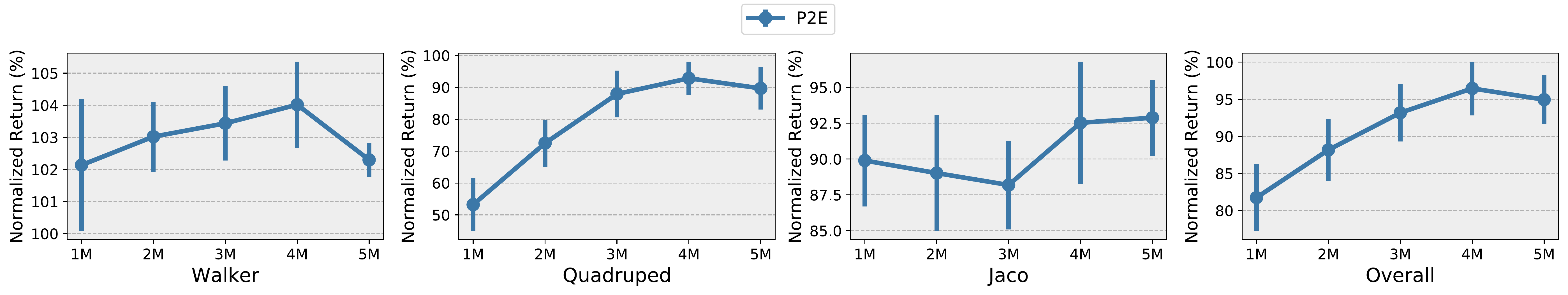}
    \includegraphics[width=.9\linewidth]{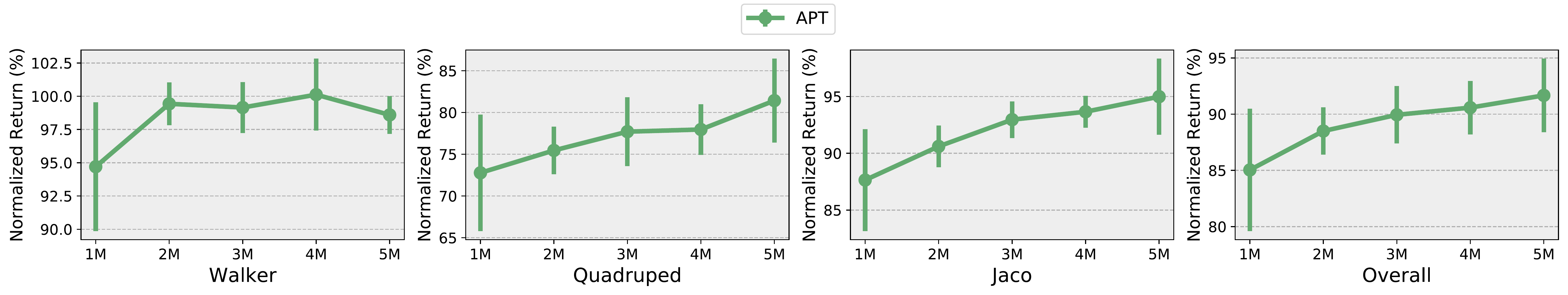}
    \includegraphics[width=.9\linewidth]{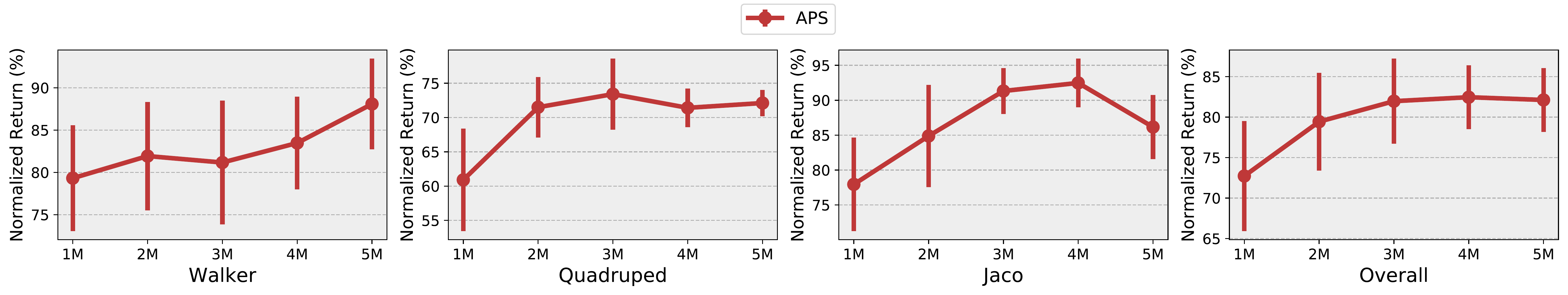}
    \caption{\textbf{Longer pre-training.} Fine-tuning performance of our method when pre-training for longer than 2M steps. Every bar reports mean performance and standard errors.}
    \label{fig:longer}
\end{figure*}

\textbf{Can a pre-training stage longer than 2M frames be beneficial?} In Figure \ref{fig:longer}, we report FT results with our full method, every 1M frames up to 5M PT frames. The aggregated results show that, adopting our method, longer PT can increase performance further, especially until 4M steps. The performance in all domains keeps increasing or remains steady until 5M steps, with two exceptional cases, Walker for Plan2Explore and Jaco for APS, where performance slightly drops between 4M and 5M steps.

For these experiments, we kept the size of the model and all the hyperparameters unvaried with respect to the 2M PT frames experiments but we increased the replay buffer maximum size to 5M frames. Increasing model capacity, and adopting additional precautions, such as annealing learning rate, it is possible that the agent could benefit even more from longer pre-training and we aim to analyse this more in details for future work.

\newpage
\section{RWRL settings}
\label{app:rwrl}

Algorithms developed in simulation struggle to transfer to real-world systems due to a series of implicit assumptions that are rarely satisfied in real environments,  e.g. URLB assumes the dynamics between PT and FT stay the same. The RWRL benchmark~\citep{DulacArnold2020RWRL} considers several challenges that are common in real-world systems and implements them on top of DMC tasks.

We take the Quadruped and Walker tasks from the RWRL benchmark and replace the low-dimensional sensor inputs with RGB camera inputs. While this removes some of the perturbations planned in the benchmark~\citep{DulacArnold2020RWRL}, such as noise in the sensors, it introduces the difficulty of a different dynamics in pixel space (due to the other perturbations), compared to the one observed during pre-training in the vanilla simulation environment.

\begin{table*}[h]
\small
\centering
\begin{tabular}{|l|llllll}
\hline
\textbf{Setting}               & \multicolumn{2}{l|}{\textbf{Easy}}    & \multicolumn{2}{l|}{\textbf{Medium}}       & \multicolumn{2}{l|}{\textbf{Hard}}        \\
\hline

\textbf{System Delays}             & \multicolumn{2}{l|}{\textit{Time Steps}}                 & \multicolumn{2}{l|}{\textit{Time Steps}}         & \multicolumn{2}{l|}{\textit{Time Steps}}                \\
Action                       & \multicolumn{2}{l|}{3}    & \multicolumn{2}{l|}{6}     & \multicolumn{2}{l|}{9}     \\ 
Rewards                      & \multicolumn{2}{l|}{10}                      & \multicolumn{2}{l|}{20}   & \multicolumn{2}{l|}{40} \\ \hline
\textbf{Action Repetition}             & \multicolumn{2}{l|}{1}                 & \multicolumn{2}{l|}{2}         & \multicolumn{2}{l|}{3}                \\ 
\hline
\textbf{Gaussian Noise}                     & \multicolumn{2}{l|}{\textit{Std. Deviation}}       & \multicolumn{2}{l|}{\textit{Std. Deviation}}   & \multicolumn{2}{l|}{\textit{Std. Deviation}}     \\ 
Action              & \multicolumn{2}{l|}{0.1}     & \multicolumn{2}{l|}{0.3} & \multicolumn{2}{l|}{1.0}         \\ \hline
\textbf{Perturbation}                  & \multicolumn{1}{l|}{\textit{[Min,Max]}}                    & \multicolumn{1}{l|}{\textit{Std.}}       & \multicolumn{1}{l|}{\textit{[Min,Max]}}                    & \multicolumn{1}{l|}{\textit{Std.}}       & \multicolumn{1}{l|}{\textit{[Min,Max]}}                     & \multicolumn{1}{l|}{\textit{Std.}}      \\ 
\textbf{Quadruped}                  & \multicolumn{1}{l|}{}                   & \multicolumn{1}{l|}{}     &      \multicolumn{1}{l|}{}               &  \multicolumn{1}{l|}{}     &                \multicolumn{1}{l|}{}     &  \multicolumn{1}{l|}{}    \\ 
(shin length)             & \multicolumn{1}{l|}{[0.25, 0.3]}                       & \multicolumn{1}{l|}{0.005} & \multicolumn{1}{l|}{[0.25, 0.8]}                       & \multicolumn{1}{l|}{0.05} & \multicolumn{1}{l|}{[0.25, 1.4]}                       & \multicolumn{1}{l|}{0.1} \\ \hline
\textbf{Perturbation}                  & \multicolumn{1}{l|}{\textit{[Min,Max]}}                    & \multicolumn{1}{l|}{\textit{Std.}}       & \multicolumn{1}{l|}{\textit{[Min,Max]}}                    & \multicolumn{1}{l|}{\textit{Std.}}       & \multicolumn{1}{l|}{\textit{[Min,Max]}}                     & \multicolumn{1}{l|}{\textit{Std.}}      \\ 
\textbf{Walker}                  & \multicolumn{1}{l|}{}                   & \multicolumn{1}{l|}{}     &      \multicolumn{1}{l|}{}               &  \multicolumn{1}{l|}{}     &                \multicolumn{1}{l|}{}     &  \multicolumn{1}{l|}{}    \\ 
(thigh length)             & \multicolumn{1}{l|}{[0.225, 0.25]}                       & \multicolumn{1}{l|}{0.002} & \multicolumn{1}{l|}{[0.225, 0.4]}                       & \multicolumn{1}{l|}{0.015} & \multicolumn{1}{l|}{[0.15, 0.55]]}                       & \multicolumn{1}{l|}{0.04} \\ \hline
\end{tabular}
\vspace{1em}
\caption{Perturbations setting for each challenge of our adapted tasks from the RWRL benchmark, in increasing levels of intensity.}
\label{tab:challengehyperparams}
\end{table*}

\section{Extended Analysis}
\label{app:additional_analysis}

We note that, to run the experiments faster, we did not use Dyna-MPC for the extended analysis. Furthermore, the Jaco tasks used slightly differ from the original ones in URLB, only in that the target to reach cannot move. This allows consistency of the reward function between PT and FT, so that a reward predictor can be trained on `reward-labelled' PT data. However, because of this change, the performance in Jaco may differ from the other main results (mainly in Figure \ref{fig:reward_ablation} and Figure \ref{fig:mpc_results}).

\subsection{Learning Rewards Online} \label{app:learning-reward-predictor}

 In Figure \ref{fig:reward_ablation} of the main text, we measure the gap in performance between pre-trained agents that have no knowledge of the reward function at the beginning of fine-tuning and agents whose reward predictor is initialized from a reward predictor learned on top of the unsupervisedly collected data (violating the URLB settings). Crucially, the agent during unsupervised PT can learn the reward predictor without affecting neither the model learning or the exploration process. To not affect the model, gradients are stopped between the reward predictor and the rest of the world model. To not affect exploration, the rewards used to train the agent's actor and critic remain the intrinsic rewards, used for exploration. 

\subsection{Zero-shot Adaptation}\label{app:zero-shot-perf}

Using agents that have access to a PT reward predictor, we explore the idea of zero-shot adaptation using MPC, which is trying to solve the URLB tasks using only planning and the pre-trained world model and reward predictor.
In order to obtain good performance, this assumes that the model correctly learned the dynamics of the environment and explored rewarding transitions that are relevant to the downstream task, during pre-training. 
In Figure \ref{fig:mpc_results} of the main text, we compare the results of performing MPC in a zero-shot setting (ZS) with the performance of an MPC agent that is allowed 100k frames for fine-tuning (FT). As for the MPC method, we employ MPPI~\citep{Williams2015MPPI}. Because these experiments are particularly expensive to run, we just them on the agents trained with the Plan2Explore URL approach. 

We observe that the performance of zero-shot MPC is generally weak. While it overall performs better than the non-pre-trained model, simply applying MPC leveraging the pre-trained world model and reward predictor trained on the pre-training stage data is not sufficient to guarantee satisfactory performance. The fact that exploiting the fine-tuning stage using the same MPC approach generally boosts performance demonstrates that the model has a major benefit from the FT stage. Still, the performance of MPC generally lacks behind the actor-critic performance, suggesting that, especially in a higher-dimensional action space such as the Quadruped one, amortizing the cost of planning with actor-critic seems crucial to achieve higher performance.

\subsection{Latent Dynamics Discrepancy}\label{app:LD-discrepancy}

 Model misspecification is a useful measure to assess the uncertainty or inaccuracy of  the model dynamics. It is computed as the difference between the dynamics predictions and the real environment dynamics. The metric helps build robust RL strategies, that take the dynamics uncertainty into account while searching for the optimal behavior~\citep{Talvitie2018RewardMiss}. However, with pixel-based inputs the dynamics of the environment are observed through high-dimensional images. And this in-turn could hurt the metric evaluation, since the distances in pixel space can be misleading. In our approach, we use a model-based RL agent that learns the dynamics model in a compact latent space $\mathcal{Z}$.
 
Our novel metric, \emph{Latent Dynamics Discrepancy} (LDD), quantifies the ``misspecification" of the learned latent dynamics accordingly. The metric quantifies the distance between the predictions of the pre-trained model and the same model after fine-tuning on a downstream task. However, as the decoder of the world model gets updated during fine-tuning, the latent space mapping between model states $z$ and environment states $s$ might drift. For this reason, we freeze the agent's decoder weights, so that the model can only improve the posterior and the dynamics. This ensures that the mapping $\mathcal{Z} \xrightarrow[]{} \mathcal{S}$ remains unchanged and allows to compare the dynamics model after fine-tuning with the one before fine-tuning. In order to measure the distance between the distribution output by the dynamics network, we chose the symmetrical Jensen-Shannon divergence:
\begin{equation}
    \textrm{LDD} = \E_{(z_t, a_t)}\big[ D_{\textrm{JS}}[p_{\textrm{FT}}(z_{t+1}|z_t, a_t) \Vert {p_\textrm{PT}}(z_{t+1}|z_t, a_t)]\big], 
\end{equation}
where the expectation is taken over the previous model states $z_t$ sampled from the fine-tuned posterior $q_\textrm{FT}(z_t)$, actions $a_{t-1}$ sampled from an oracle actor $\pi^*(a_t|z_t)$, so that we evaluate the metric on optimal trajectories, whose environment's state distribution corresponds to the stationary distribution induced by the actor $s_t \sim d^{\pi^{*}}(s_t)$. We used 30 trajectories per task in our evaluation.

We observe in our experiments that  there exists a correlation between the metric and the performance ratio between a zero-shot model and a fine-tuned model (see Figure \ref{fig:model_miss} in the main paper).  The key observation is that major updates in the model dynamics during fine-tuning phase played an important role in improving the agent's performance, compared to the pre-trained model and zero-shot performance. 
Future research may attempt to reduce such dependency by either improving the model learning process, so that the pre-trained dynamics could have greater accuracy, or the data collection process, proposing URL methods that directly aid to reduce such uncertainty.

\subsection{Unsupervised Rewards and Performance}\label{app:unsup-rew-perf}
We further analyzed the correlation between the normalized performance of the different exploration agents and their intrinsic rewards for optimal trajectories obtained by an oracle agent. A strong negative correlation between the two factors should indicate that the agent is more interested in seeing the optimal trajectories when its performance is low on the task. 

We observe that there is negative correlation between Plan2Explore~(P2E), ICM, LBS's performance and their intrinsic rewards, while we found $\sim$0 correlation for RND (see Table \ref{tab:novelty} in the main text). Out of the methods tested, LBS significantly demonstrated the correlation, as its p-value is $<0.05$. This is likely one of the key factors for the high performance of the agent using LBS on the benchmark.

One possible explanation is that LBS searches for transitions of the environment that are difficult to predict for the dynamics, so the model likely learns those transitions more accurately, facilitating planning during the fine-tuning stage. Another potential explanation is that, given the high correlation between intrinsic and extrinsic rewards, the actor initialized by LBS performs better at the beginning of FT, speeding up adaptation.

\newpage

\section{On the sparsity of the Jaco tasks}

\begin{figure}
    \centering
    \includegraphics[width=0.6\textwidth]{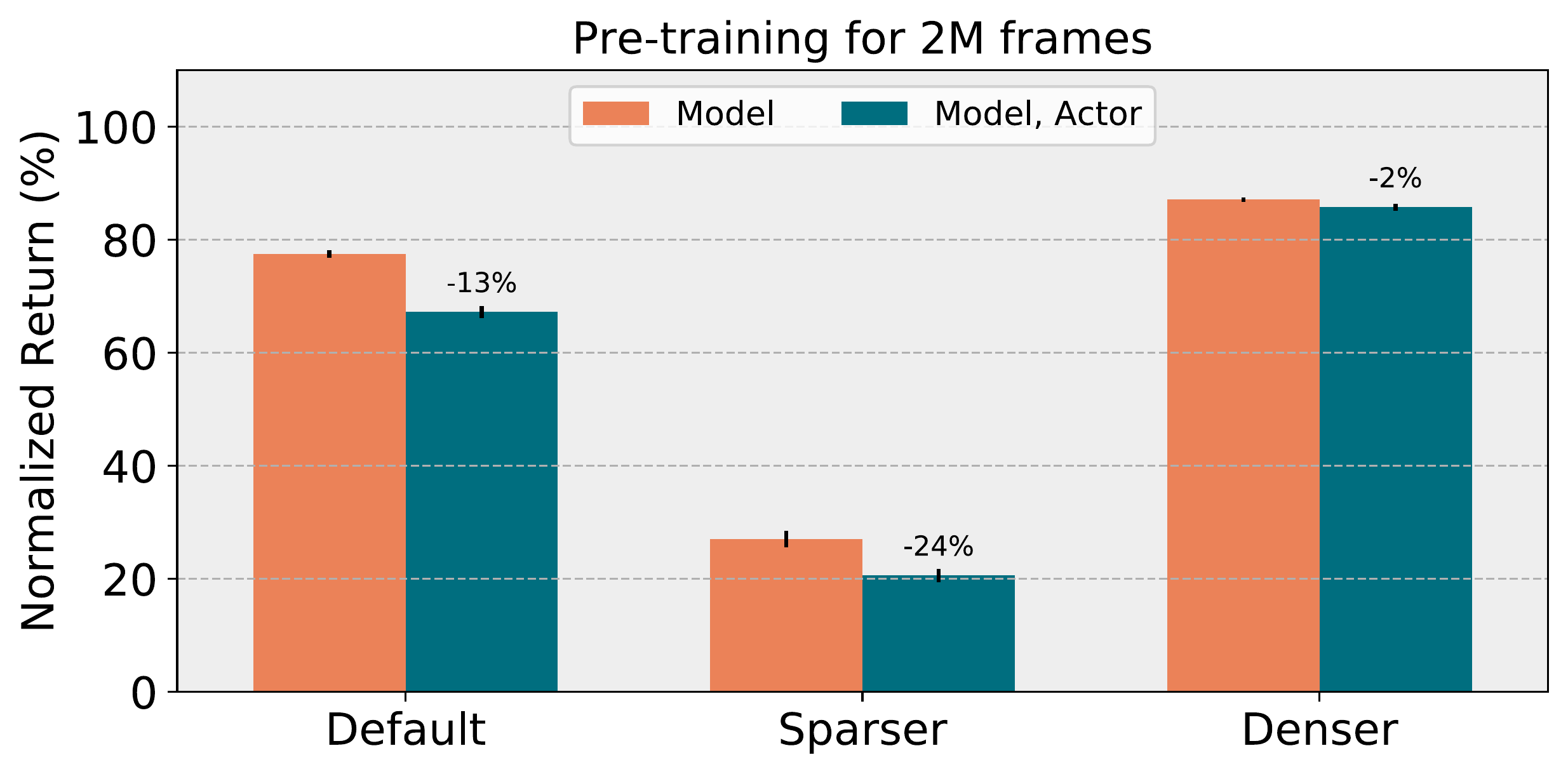}
    \caption{\textbf{Jaco tasks variations.} Testing denser and sparser rewards for the Jaco tasks. The performance gap between choosing to use or not to use the pre-trained actor increases with the sparsity of the reward function.}
    \label{fig:sparse_jaco}
\end{figure}

The Jaco tasks in URLB are sparsely rewarded reaching tasks. To support our claim that fine-tuning the exploration actor is challenging due to the task sparsity, we conducted additional experiments on two modified versions of the Jaco tasks: one with a sparser reward function and another with a denser reward function. In these experiments, the maximum reward obtainable is the same as the default URLB tasks, but the area of the environment that is rewarded is smaller in the sparser version and larger in the denser version. 

Results are presented in Figure \ref{fig:sparse_jaco} (averaged across all 2M steps PT methods and all Jaco tasks). Our findings indicate that the sparser the reward function, the greater the performance difference between initializing with the exploration PT actor (Model, Actor) and a random actor (Model). Based on visual inspection of the agent's behavior (which can be seen on the \href{https://masteringurlb.github.io/}{project website}), we believe this occurs because exploration policies often move the agent toward areas far from the initial position, which are typically further away from the target and make the reward harder to find compared to a random policy, which tends to explore closer to the initial pose.

\section{A recipe for unsupervised RL}

In our large-scale study, we explored several design choices to establish the most adequate approach to tackle the URL benchmark, aiming to provide a general recipe for data-efficient adaptation thanks to unsupervised RL. Three main findings about useful strategies to apply for URL emerge from our study:
\begin{enumerate}
    \itemsep0em 
    \item \textit{unsupervised model-based PT}: learning a model-based agent with data collected using unsupervised RL (Figure \ref{fig:expl_results});
    \item \textit{performing task-aware FT}: fine-tuning the PT world model (always) and the pre-trained actor (where beneficial), while learning the critic from scratch (Figure \ref{fig:modules_results});
    \item \textit{using a hybrid planner}: such as Dyna-MPC, to further improve data efficiency (Figure \ref{fig:planning_improv}).
\end{enumerate}
An overview of our method is illustrated in Figure~\ref{fig:overview} and a detailed algorithm is presented in Appendix~\ref{app:algorithm}.


We believe the above recipe could be applied to several unsupervised settings, outside of URLB, with the precaution that one should pay attention to two aspects: (a) whether starting fine-tuning from the PT actor is meaningful for the downstream task, (b) what is the best data collection strategy to adopt in the adopted domain. 

\newpage
\section{Hyperparameters}
\label{app:hyperparams}

Most of the hyperparameters we used for world-model training are the same as in the original DreamerV2 work~\citep{Hafner2021DreamerV2}. Specific details are as outlined here:

\begin{table}[h!]
\centering
\begin{tabular}{lc}
\toprule
\textbf{Name} & \textbf{Value} \\
\midrule
World Model \\
\midrule
Batch size & 50 \\
Sequence length & 50 \\
Discrete latent state dimension & 32 \\
Discrete latent classes & 32 \\
GRU cell dimension & 200 \\
KL free nats & 1 \\
KL balancing & 0.8 \\
Adam learning rate & $3\cdot10^{-4}$ \\
Slow critic update interval & 100 \\
\midrule
Actor-Critic \\
\midrule
Imagination horizon & 15 \\
$\gamma$ parameter & 0.99 \\
$\lambda$ parameter  & 0.95 \\
Adam learning rate & $8\cdot10^{-5}$ \\
Actor entropy loss scale & $1\cdot10^{-4}$ \\
\midrule
Dyna-MPC \\
\midrule
Iterations & 12 \\
Number of samples & 512 \\
Top-k & 64 \\ 
Mixture coefficient (Actor/CEM) & 0.05 \\
Min std (fixed) & 0.1 \\
Temperature & 0.5 \\ 
Momentum & 0.1 \\ 
Planning horizon & 5\\ 
\midrule
Common \\
\midrule
Environment frames/update & 10 \\
MLP number of layers & 4 \\
MLP number of units & 400 \\
Hidden layers dimension & 400 \\
Adam epsilon & $1\cdot10^{-5}$ \\
Weight decay & $1\cdot10^{-6}$ \\
Gradient clipping & 100 \\
\bottomrule
\end{tabular}
\vspace{1em}
\caption{World model, actor-critic, planner (Dyna-MPC) and common hyperparameters.}
\label{tab:hparams}
\end{table}

For the pure MPC-based experiments, we increased the number of MPPI samples from 512 to 1000, the number of top-k from 64 to 100, and the horizon from 5 to 15, to compensate for the absence of the actor network's samples and the critic's predictions in the return estimates. 

\end{document}